\documentclass[3p,preprint]{elsarticle}
\usepackage{amsmath,amssymb,amsfonts,amsthm}
\usepackage[T1]{fontenc}
\usepackage[OT1]{fontenc} 
\usepackage{algorithm}
\usepackage{algpseudocode}
\usepackage{algcompatible}
\usepackage{array}
\usepackage{textcomp}
\usepackage{booktabs}
\usepackage{stfloats}
\usepackage{url}
\usepackage{verbatim}
\usepackage{graphicx}
\usepackage{epsfig}
\usepackage{subfigure}
\usepackage{multirow}
\usepackage{multicol}
\usepackage{enumitem}
\usepackage{lipsum}
\usepackage{lineno}
\usepackage{orcidlink}
\graphicspath{{./FIG/}}
\hypersetup{pdfauthor=author}

\DeclareGraphicsExtensions{.pdf,.jpeg, .jpg, .png}
\def\BibTeX{{\rm B\kern-.05em{\sc i\kern-.025em b}\kern-.08em
    T\kern-.1667em\lower.7ex\hbox{E}\kern-.125emX}}
\usepackage{balance}

\usepackage{makecell}
\usepackage{threeparttable} 

\graphicspath{{./FIG/}}
\DeclareMathOperator*{\argmin}{argmin}

\journal{Computer Speech and Language}

\begin{document}


\title{Large Language Model Pruning}

\author[add1,add2]{Hanjuan~Huang~\orcidlink{0009-0006-4691-8062}}\ead{huanghanjuan66@gmail.com}
\author[add1]{Hao-Jia Song~\orcidlink{0009-0007-9633-1312}}\ead{hhh5911@gmail.com}

\author[add1]{Hsing-Kuo Pao~\orcidlink{https://orcid.org/0000-0002-5518-9475}\corref{cor1}}\ead{pao@mail.ntust.edu.tw}\cortext[cor1]
{Corresponding author}

\address[add1]{Dept. of Computer Science and Information Engineering\\
National Taiwan University of Science and Technology, Taipei, Taiwan 106}

\address[add2]{College of Mechanical and Electrical Engineering,\\
WUYI University, Wuyishan, 354300, China}


\begin{abstract}
    We surely enjoy the larger the better models for their superior performance
    in the last couple of years when both the hardware and software support
    the birth of such extremely huge models.
    The applied fields include text mining and others.
    In particular, the success of LLMs on text understanding
    and text generation draws attention from researchers
    who have worked on NLP and related areas for years or even decades.
    On the side, LLMs may suffer from problems like model overfitting,
    hallucination, and device limitation to name a few. In this work,
    we suggest a model pruning technique specifically focused on LLMs.
    The proposed methodology emphasizes the explainability of deep learning models.
    By having the theoretical foundation, we obtain a trustworthy deep model so that
    huge models with a massive number of model parameters become not quite necessary.
    A mutual information-based estimation is adopted to find neurons
    with redundancy to eliminate. Moreover, an estimator with well-tuned
    parameters helps to find precise estimation to guide the pruning procedure.
    At the same time, we also explore the difference between pruning on
    large-scale models vs. pruning on small-scale models.
    The choice of pruning criteria is sensitive in small models
    but not for large-scale models. It is a novel finding through this work.
    Overall, we demonstrate the superiority of the proposed model
    to the state-of-the-art models.
\end{abstract}




%


\maketitle

\section{Introduction}
\label{sec:intro}

People embrace large-scale models based on large-scale data nowadays.
The bigger the models, the better performance we can expect
from the models. This is because the big models can have
their complexity large enough to cover the high diversity that
could be seen from the data.
Given this reason, giant industries look for Large Language Models (LLMs)
starting from a few years back, and it gets heated since
the announcement of ChatGPT at the end of 2022.
That shows the evidence that people generally believe the larger the models
the more problems can be solved by the models.

The progress of LLMs can be seen by various well-known models such as
LLaMA2~\cite{touvron2023llama}, PaLM-2~\cite{anil2023palm}, and their followers for
various scenarios given a diverse set of data such as LLaMA-Pro-8B~\cite{wu2024llama}
and Med-PaLM 2~\cite{singhal2023towards}. At the same time, some others are focused on
proposing relatively small models to achieve the same model effectiveness,
such as TinyLlama-1.1B~\cite{zhang2024tinyllama} and ORCA-2~\cite{mitra2023orca}.
Other threads of development includes the Gemini-based structures,
such as Gemini 1.5~\cite{reid2024gemini} and Gemma~\cite{team2024gemma};
also, InternLM2~\cite{zheng2024InternLM2} from an independent research group,
not to forget the still close-to-the-top-performance one, GPT-4~\cite{achiam2023gpt}
which top others on various scenarios and receive quite a rich set of comments
from the general public.

LLMs are welcome if we do not have storage or computation limitations.
Therefore, applications like edge computing cannot be the target
when LLMs are among the essential parts of the system.
But even if we do not have the limitations, we should keep the storage or computation
constrained in as small size as possible for the sack of green concerns.
As an alternative viewpoint, we can ask for small models
to avoid the possibility of overfitting and hallucination. The smaller the models are,
the less likely the models can run into those cases.

In this work, we propose a pruning method that is designed especially for LLMs.
The strategy is based on some understanding of the models,
such as exploring the direction of model explainability borrowing
the Tishby's deep learning theory~\cite{tishby2000information,
tishby2015deep, shwartz2017opening}.
With the understanding of the neurons' content from the representation layer,
we have clues to prune the model where the neurons with redundancy shall be dropped
and the neurons with essential information can be kept.

The pruning on large and small-scale models can face very different situations.
In large-scale models, neurons are considered with a few possible substitutions.
That is, we usually have an over-complete representation in a large-scale model
with more than necessary neurons in the representation layer.
In this case, two or more neurons could own similar meanings or importance
and the pruning on either one can produce resulting models with similar effectiveness.
On the other hand, pruning on small-scale models can be crucial
in the sense that the number of neurons is close to the minimum required
number of neurons and pruning on such models needs careful treatment
with as precise computation as possible.
In this work, the proposed pruning method, with the help of Mutual Information (MI) estimation
can serve the need for such precise computation that leads to
effective compressed models.




The most recent progress in LLM compression includes direct network pruning,
knowledge distillation, quantization, and low-rank factorization.
The different methods enforce the compression on different network parts.
Direct network pruning~\cite{lecun1989optimal, han2015learning,li2016pruning,
ma2023llm, frantar2023massive} is aimed to remove redundant or unimportant components (neurons or weights)
to have a small network to replace a large network
to fulfill the goal. The knowledge distillation~\cite{hinton2015distilling, tung2019similarity, gu2023knowledge}
is focused on finding a small, student model to learn what can be offered from a large, teacher model.
Mimicking the prediction ability of the large model from a small model,
we achieve the goal of model downsizing.
Quantization~\cite{liu2021post, liu2023llm, ding20224} provides another approach which saves space
by utilizing integers or discrete numbers to substitute floating-point numbers in networks.
The computation time may also be saved by such a design.
The last approach, called low-rank factorization~\cite{povey2018semi, wu2023zeroquant},
emphasizes replacing the large weight matrix with a small weight matrix to reach the goal of model compression.

To focus on the family of network pruning techniques, we consider two types of methods,
the structured and unstructured ones. The overall goal is to remove redundant components
from the models. First, the structured pruning simplifies an LLM by
removing the entire structural components, such as neurons, channels, or layers while keeping
the network structures~\cite{blalock2020state, zhu2023survey}.
On the other hand, the unstructured pruning~\cite{frantar2023massive, zhang2023pruning}
is aimed at pruning the redundant neurons or links. Given the approach of deleting individual parameters,
we may suffer from the irregular sparse structure problems.
To compare between the two, the structured pruning
can be deployed to various edge devices directly, while the unstructured pruning
needs to be accompanied by the assistance of extra software or hardware treatment
to complete the task~\cite{zhu2023survey}.

We have to keep in mind a few items when designing an appropriate compression
method for LLMs. First, quite a few LLMs were trained mainly based on unlabeled data.
Some of these types of pre-trained LLMs can be used by third-party developers
for various downstream tasks. Therefore, we do not know and we cannot
assume any label or contextual information provided when building the LLMs.
That is to say, we do not know what could be the real mission
when building the models or applying any compression to
the models~\cite{nova2023gradient}.
After all, we better assume only an unsupervised type of compression applied to LLMs.
Second, most pruning methods need retraining
to maintain the models' performance. This retraining procedure
inevitably needs to consume various resources~\cite{nova2023gradient}
and should be avoided if possible.
In LLMs, the Feed-forward Network (FFN) is the major part of the models
in terms of the models' size and it attracts
the most computation~\cite{ganesh2021compressing}.
Therefore compressing this part of a network means a lot if refers to
the saving of space and time complexity. All the above points motivate us to search
for an unsupervised LLM compression method that needs no further retraining after
the compression is done. Moreover, we prefer one pruning method
that can be applied to FFN in particular and that is structured typed.

We utilize an information-based estimation method to measure the information
stored in network neurons to decide how we can operate the compression procedure.
An MI-based estimator helps to find the relation between
two groups of random variables. Both the linear and nonlinear
relations between random variables can be found by
such MI estimation~\cite{cover1999elements}.
In this work, we estimate the MI value to decide
how related between two random variables that correspond to two
hidden neurons. When two neurons own a high MI value,
they share overlapped information and one of them can be pruned
without too much information loss and the overall procedure of model pruning
is operated similarly to produce small-scale models in the end.


The aforementioned idea is simple to understand but not trivial
to operate in reality due to the target of this work is to focus
on the pruning of LLMs. In general, dealing with the pruning
of large-scale deep networks is hard or intractable
when the model size or the dimensionality in the hidden layers
is beyond a certain scale. The MI estimation cannot be scalable easily
and we have to rely on an effective MI estimation
when facing high-dimensional multivariate data
or hidden layers that own large-scale neurons.
Given the studies from
Wickstr{\o}m et al.~\cite{wickstrom2019information},
Giraldo et al.~\cite{giraldo2014measures}, and
Yu et al.~\cite{yu2019multivariate}, researchers
aware that the R\'enyi's $\alpha$-order entropy estimator
can help to deal with high-dimensional scenarios.
Specifically, a matrix-based R\'enyi's $\alpha$-order entropy estimator 
can be used to estimate R\'enyi's entropy in FC layers while
a tensor-based R\'enyi's $\alpha$-order entropy estimator
can be used to estimate the R\'enyi's entropy in the CNN layers.
In this work, we adopt the matrix-based
R\'enyi's $\alpha$-order entropy estimator to realize
the MI computation and proceed with the pruning algorithm
to find a small but similar effectiveness model
known as the model pruning given the LLMs.


To estimate the MI between hidden neurons as precisely as possible,
one can rely on the result offered by R\'enyi's $\alpha$-order entropy estimator.
We point out that a parameter in the estimation, the kernel width parameter $\sigma$
is a key to having precise estimation results.
Moreover, in this work, we propose a novel approach to choose
an appropriate parameter $\sigma$,
therefore to have a better R\'enyi's entropy estimation,
and can accurately prune the redundant neurons
to build a small, but effective deep model in the end.



We summarize the proposed work and its contributions as follows.
\begin{itemize}
    \item The proposed method is considered an unsupervised approach,
    which needs no label information to decide the pruning strategy.
    That gives us less burden when moving to large-scale models
    which may suffer from the labeled data hungry problem.
    \item The proposed method is a structural pruning method with easy implementation network-wise.
    \item No need to retrain or fine-tune the compressed model from the proposed method.
    That implies the pruning strategy can indeed target the neurons
    that are redundant. On the other hand, if there needs heavy fine-tuning
    after the compression to have an effective model, the compression is considered
    not able to pinpoint the redundant neurons or no clear redundant neurons
    can be found. Such a weak compression only confirms the correctness of the number of neurons
    but not the content of the neurons before the compression.
    \item The method has its superiority over other unsupervised pruning methods.
    The method also shows some competitiveness even if compared to some
    of the supervised approaches.
    \item After all, a new kernel width estimation method is introduced
    to compute the MI between hidden nodes.
\end{itemize}

The rest of the paper is organized as follows. We elaborate on the background knowledge that
is related to model pruning and LLMs in Sec.~\ref{sec:past},
which is followed by a detailed explanation of the proposed model pruning methodology
and the estimation techniques in Sec.~\ref{sec:method}.
To evaluate the proposed method, we have all the results
and discussion presented in Sec.~\ref{sec:exp} and in Sec.~\ref{sec:conc} we conclude this work.
\section{Past Work}
\label{sec:past}

Let us discuss the past work that is related to deep model pruning,
especially the pruning which can be applied to LLMs.
At the same time, we should keep an eye on the most recent progress of LLMs,
and understand what pruning techniques can be used for LLM pruning,
given the most recent popular model structures to focus for the time being.
To speak of the proposed pruning method,
we also need to mention some different types of pruning methods,
as well as the MI estimation which is the metric
that we use to guide the pruning procedure.

\subsection{Structured Pruning for LLMs}
Let us go through the past pruning methods that address large-scale models
or LLMs and explain how they can achieve the goal.
If we consider the structured pruning for LLMs, we separate the methods into
the supervised and unsupervised types. In supervised methods,
we focus on the relation between hidden neurons and the label,
while the mutual relation between different groups of neurons receives
more attention in the unsupervised methods. Which type of pruning should be used
also depends on the applications.

First, we start with the supervised pruning methods, which are also
the majority in the LLM pruning up to
now~\cite{voita2019analyzing, liu2021ebert, 
kwon2022fast, yang-etal-2022-textpruner, park2023accurate, ma2024llm}.
Among them, Voita et al.~\cite{voita2019analyzing} proposed a pruning method
based on stochastic gates and a differentiable relaxation of $L_0$ penalty,
which can remove the vast majority of heads without seriously
affecting the model's performance.
Liu et al.~\cite{liu2021ebert} proposed a structured pruning method for
efficient BERT inference (EBERT), which can dynamically prune unimportant heads
in Multi-Head Self-attention (MHA) and unimportant channels in FFN
with the help of the predictor branch.
That means that the labeled data are necessary for their operations.
Kwon et al.~\cite{kwon2022fast} proposed a three-stages pruning framework,
which used a Fisher-based mask search algorithm (labeled data are needed)
to decide which heads/filters to prune,
then rearranged the pruned heads/filters, and at last tuned the mask variables
to recover the output signal for each layer.
Yang et al.~\cite{yang-etal-2022-textpruner} proposed
a model pruning toolkit called TextPruner
for pre-trained language models. The toolkit includes
two pruning methods: one is supervised 
method, which used the training loss to measure the importance score of neurons;
and the other is the self-supervised method, which used
the Kullback-Leibler divergence
to measure the importance score of neurons. Park et al.~\cite{park2023accurate}
proposed a structured pruning algorithm, named Kprune (Knowledge-preserving),
which focused on preserving the useful knowledge of the pre-trained model
to minimize pruning errors through an iterative pruning process that consisted of
knowledge measurement, knowledge-preserving mask search,
and knowledge-preserving weight-tuning. Ma et al.~\cite{ma2024llm}
introduced an LLM pruning approach
referred to as LLM-Pruner. This method employed structural pruning,
selectively eliminating non-essential coupled structures guided
by gradient information.
The aim is to preserve the majority of the LLM's functionality
to the fullest extent possible.

Some of the discussion on the structured and unstructured pruning can be found in
the Sec.~\ref{sec:intro}. Other than that,
we point out one of the essential procedures,
performing retraining or not after pruning, and exploring this viewpoint
through various prior work. The work belonging to
the retraining-based group~\cite{liu2021ebert, lagunas2021block, ma2024llm}
has the entire dataset ready for retraining on the compressed model.
On the other hand, the retraining-free
methods~\cite{kwon2022fast, park2023accurate, an2023fluctuation}
can skip this procedure as the model after pruning remains to be effective.
In general, training or retraining on LLMs is known to be resource-intensive
and retraining-free approaches should be favored in this sense.
An et al.~\cite{an2023fluctuation} and Park et al.~\cite{park2023accurate}
saved the retraining step, but found some alternative ways
to enhance the model's effectiveness.
For instance, An et al.~\cite{an2023fluctuation} added additional bias terms
to recover the output feature maps using the baseline values.
Park et al.~\cite{park2023accurate} tuned the weights on the pruned model
for model improvement. Also, Kwon et al.~\cite{kwon2022fast} asked
for label information to help with model improvement even
if retraining after pruning is not needed.

Overall, we understand that most LLMs have the models trained without the help of label information.
The label information for most LLM tasks is indeed hard to acquire.
Therefore, we rely on unsupervised pruning more than before.
Nova et al.~\cite{nova2023gradient} proposed a 
gradient-free structured pruning framework to integrate two ranking techniques
representative ranking and data-driven ranking, without the help of labeled data

\subsection{Feature Selection based on Mutual Information }

In a sense, we can recognize the values of all network neurons as feature content.
Pruning neurons is equivalent to feature selection in this case.
In terms of feature selection, we aim to remove
irrelevant features~\cite{zhou2022feature} from the set
to shrink the model size. The irrelevant features imply two meanings: the ones
may own redundancy or the ones help nothing
in the focused prediction task from the model.
We call the first one the unsupervised type and the second one the supervised type
and further discuss the pruning strategies of the two types.

If through the MI~\cite{cover1999elements} estimation between different neurons,
we may understand the neurons' relation and use the clue
to prune the possible non-important neurons from the network.
As mentioned before, we can utilize MI to measure the relation 
either between two hidden neurons or between a hidden neuron
and the target information.
The first is called the measurement of redundancy
and the second is called the measurement of relevancy~\cite{zhou2022feature}

Lewis~\cite{lewis1992feature} is the first one to propose
the MI-based feature selection method,
estimating the MI between a potential feature (or a neuron) and the label,
we eliminate the neurons that have the MI result fall below a pre-defined threshold.
Peng et al.~\cite{peng2005feature} proposed a criterion called
minimal-redundancy-maximal-relevance (mRMR) to select useful features,
which checked both the MI between candidate features and the label
and the MI between different neurons to decide what could be the features to keep.
Apparently, we have to pay attention to the trade-off
between redundancy and relevance
at the same time to make a final decision, which could be case by case.
For instance, in a transfer learning type of scenario,
we may focus more on the relevance
more because the distribution of the target task could be different from
what we can focus on the source task. Zhou et al.~\cite{zhou2022feature}
proposed another approach called CCMI to integrate MI and correlation coefficient
to measure how different features are related. Similarly, they also considered
both the redundancy and relevance at the same time such as
picking up features with high relevance to the label and low redundancy
between different features or hidden neurons.
Fan et al.~\cite{fan2021layer} applied the MI-based computation
to feature or neuron selection then to model pruning.
The redundant neurons got eliminated if the neuron set at layer $L$ after the elimination
owns the biggest MI value to the neuron set at layer $L+1$.
In the proposed method, we pay attention to redundancy more than relevance.
Moreover, we care about the different situations when a small or a large set of neurons
is considered for pruning. Other than the aforementioned issues, we utilize an alternative way
to estimate the MI, which provides more precise pruning than prior work.

\subsection{Mutual Information Estimation}

The MI estimation on deep learning networks is hard,
if not intractable due to the large-scale of network structures and data size.
The classical binning-based estimator~\cite{shwartz2017opening} considered
quantizing neurons' output to estimate the corresponding probability distribution,
which leads to at least three problems: (1) an appropriate decision on the
bin size to ensure the estimation precision~\cite{sarvani2022hrel, lee2021information},
(2) probability distribution estimation needs a large amount of samples~\cite{lee2021information},
(3) difficult computation for certain activation functions such as ReLU.
Some other issues include the possible systematic errors~\cite{holmes2019estimation}
happened in the computation procedure.
We may utilize some hyperbolic functions (e.g., $\tanh$) to deal with the last issue.
Kraskov et al.~\cite{kraskov2004estimating} proposed
a $k$NN distance-based MI estimation called KSG,
to deal with a wide range of activation functions.
However, this could rely on a wise decision on the number of neighbors.
To deal with other issues, Belghazi et al.~\cite{belghazi2018mutual} proposed
Mutual Information Neural Estimator (MINE), adopting a different network,
using the gradient descent to realize the MI estimation given high-dimensional random variables.
Through their framework, both the dimensionality of neurons and the number of samples
can be extended linearly for better estimation precision.
However, the method is sensitive to the choice of network,
and the converging speed is slow for such a network.
Wickstr{\o}m et al.~\cite{wickstrom2019information} improved from
the result of Giraldo et al.~\cite{giraldo2014measures} and Yu et al.~\cite{yu2019multivariate}
and proposed a novel matrix or tensor-based estimation called
R\'enyi $\alpha$-order entropy estimator, can estimate MI given
high-dimensional multivariate data without estimating the probability of random variables
that are involved in the MI computation.

The key to the R\'enyi $\alpha$-order entropy estimator is focusing on
the estimation of the kernel width parameter. Based on that,
we have choices between the supervised learning
and unsupervised learning approaches.
In supervised learning, an optimal criterion~\cite{wickstrom2019information}
is used, while the Scott's rule~\cite{scott2015multivariate}
is considered for the unsupervised learning case.
In a nutshell, they aligned the label kernel matrix and a kernel matrix
from a pre-specified layer to approximate the kernel width parameter.
In Scott's rule~\cite{scott2015multivariate}, one can estimate
the kernel width parameter by checking the data size and the dimensionality
of the focused hidden layer.
The current approach to applying Scott's rule
is focused on the estimation of the whole hidden layer~\cite{lee2021information}.

\section{Proposed Method}
\label{sec:method}

In this section, we introduce the proposed method in detail.
The goal of this work is to have an explainable model
and prune the model based on such explainability.
The key steps to achieve the goal include:
finding a way to confirm the explainability of the model,
taking advantage of the model's explainability to prune neurons
in the representation layer based on MI computation,
and having an effective estimation of MI
to produce the final result.
Before we go on to elaborate on the details of the proposed method,
we introduce the notations that shall be used in this work.

\subsection{Notations}
\label{subsec:notations}

In the $\ell$-th transformer encoder, 
we have $K_\ell$ neurons in the fully-connected layer
of FFN, which are denoted by ${\cal Z}_1,{\cal Z}_2,\ldots,{\cal Z}_{K_\ell}$
and we use the random variables $Z_1,Z_2,\ldots,Z_{K_\ell}$
to describe the value of those neurons (features) in the FC layer.
That is, ${\cal Z}$ refers to a neuron and $Z$ is the random variable
to describe the value on neuron ${\cal Z}$.

\subsection{Framework}

\begin{figure}[H]
\centering 
\includegraphics[width=0.8\textwidth]{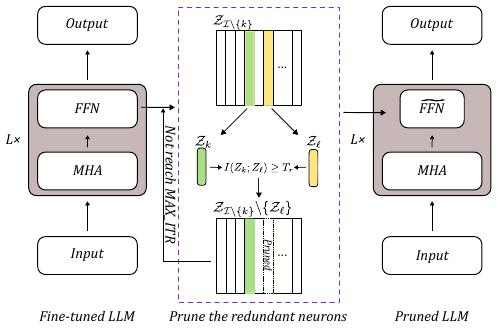}
\caption{The pruning procedure has the input LLM with $L$ transformer blocks
and produces the resulting model with the pruned FFN FC layer.
We randomly select two neurons $k$ and $\ell$ and compute their MI $I(Z_k;Z_\ell)$.
If their MI value is smaller than a pre-specified threshold $T_r$,
one of the neurons is deleted. We repeat the step until a maximum number of iterations is reached. 
}
\label{fig:fc}    
\end{figure}

Given a pre-trained LLM, we perform a fine-tuning procedure
that is designed for a specific task,
then apply the proposed pruning to the fine-tuned model
to obtain its compressed version.
The compressed model is assumed to have similar behavior
to its original model.
The overall proposed methodology is shown in Fig.~\ref{fig:fc}

\subsection{The Proposed Pruning Methodology}

\subsubsection{Working on Feed-forward Networks}

In one typical design of transformers,
we have Multi-Head Self-attention (MHA), followed by one Feed-forward Network
in its encoder part. Moreover, each FFN contains two linear transformation layers
and a GeLU activation in between.

\subsubsection{Redundancy as Feature Selection Criteria}

We adopt {\em mutual information} to measure the relationship between features.
Based on the result, we prune features with a certain level of redundancy.
In the fully connected layer of FFN,
we randomly select two features represented by their corresponding
random variables $Z_k$ and $Z_\ell$,
and compute their mutual information as $I(Z_k; Z_\ell)$. 
If the values of $I(Z_k; Z_\ell)$ is large enough
to show a certain degree of information overlap,
we choose one to delete from the feature set.

The whole procedure of the pruning strategy is shown
in Algorithm~\ref{alg:filter}.

\begin{algorithm}[H]
\caption{The algorithm of an alternative strategy to select a subset of features
that has low mutual information between pairwise features.}
\label{alg:filter}
	\begin{algorithmic}[1]
    \REQUIRE
      \STATEx $K_{r}$: The no. of remaining features after the alternative pruning strategy
      \STATEx $T_r$: Maximum allowed feature overlapping
      \STATEx $MAX\_ITR$: The maximum number of iterations
    \ENSURE
     \STATEx ${\cal Z}_{{\cal I}\backslash\{k\}}$: The resulting feature set after the pruning strategy 
     \STATE ${\cal Z}_{{\cal I}\backslash\{k\}} \gets {\cal Z}$ \;
     \STATE $K_r = K$ \;
     \FOR {$i=1,\ldots,MAX\_ITR$}
        \STATE Randomly choose two features ${\cal Z}_k$ and ${\cal Z}_\ell$ from ${\cal Z}_{{\cal I}\backslash\{k\}}$
        with their content described by
        $Z_k$ and $Z_\ell$ \;
        \STATE Calculate the mutual information $I(Z_k; Z_\ell)$ \;
	    \IF{ $I(Z_k;Z_\ell)\ge T_r$  }  
            \STATE ${\cal Z}_{{\cal I}\backslash\{k\}} \gets {\cal Z}_{{\cal I}\backslash\{k\}}\backslash \{{\cal Z}_\ell\}$
            \STATEx // Turn off one of the similar features
            \STATE {\bf decrease} $K_r$ \;
        \ENDIF
     \ENDFOR
	\end{algorithmic}  
\end{algorithm}

\subsubsection{Clustering Strategy as a Scaling up Option}
\label{subsec:mds}

The pruning algorithm, as shown in Algorithm~\ref{alg:filter} may not scale well
to a large set of neurons or goes quadratically in terms of the number of neurons
in its computation. To bypass such a burden, we consider a clustering-based procedure
to perform the pruning in a group-based manner.
In detail, we cluster features based on their similarity
while features with high similarity should go together.
Before that, we decide the number of clusters according
to different choices of compression rates.
When the clustering result is confirmed,
we choose one feature, which could be the one closest to each cluster centroid
to be the one to keep while all the other neurons of the same cluster
should be eliminated after the procedure.

The mutual information is used to decide a metric
for the clustering procedure.
Given the pairwise distances, we utilize
the multidimensional scaling (MDS)~\cite{cox2000multidimensional}
to find coordinates in a pre-specified dimensionality.
In detail, given two features ${\cal Z}_k$ and ${\cal Z}_\ell$
we compute their mutual information $I(Z_k; Z_\ell)$
and a set of pairwise mutual information is transformed into pairwise distances.
Afterward, Eq.~\ref{eq:mds} helps to find coordinates
given a pre-specified space of certain dimensionality. 
%
\begin{equation}\label{eq:mds}
    d({\cal Z}_k,{\cal Z}_\ell)=A \exp(-I(Z_k; Z_\ell)), 
\end{equation}
%
where $A$ is a constant in case if we prefer a distance between $0$ and $1$.
In the formula, the larger the mutual information between ${\cal Z}_k$
and ${\cal Z}_\ell$, the smaller the value $d({\cal Z}_k,{\cal Z}_\ell)$ is.
In the MDS-projected space, two close-by features ${\cal Z}_k$ and ${\cal Z}_\ell$
implies that they share larger mutual information $I(Z_k; Z_\ell)$.
Moreover, two features with their large mutual information $I(Z_k; Z_\ell)\ge T_r$
may end up in the same cluster and the pruning strategy in Algorithm~\ref{alg:filter}
could suggest the removal of one of the features in the pair.
That is, we group features into a cluster if they own large mutual information.
After that, only one feature per cluster is used to be the representative
once we obtain the grouping result.
In the end, we have features in different clusters if the features
have their pairwise mutual information smaller than a threshold $T_r$.

\subsubsection{Subsidiary Condition}

The procedure may not produce a unique compression model
because the solution to MDS and the selection of representatives
may not always be the same. We may suggest some side condition
to encourage a decent compression result
by trying $M$ random seeds and choosing the best one out of them
by the following criteria.


The subsidiary condition is to minimize the difference between 
the original and the compressed models.
We use the Kullback-Leibler (KL) divergence to measure the difference
between the original and the compressed model by Eq.~\ref{eq:kl}
if focusing on the representation of both models.
Given the original model, we aim to find a compressed model ${\cal M}$
that is closet to the original model in its representation distribution
if measured by KL divergence.
\begin{equation}
\label{eq:kl}
{\cal M}_{\rm comp}^* =
\argmin_{\cal M}
D_{\rm KL}(p(z_o)||p(z_{\cal M})) \, ,
\end{equation}
where $D_{\rm KL}(p_1\| p_2)$ measures the KL divergence between two distributions
$p_1$ and $p_2$, $z_o$ and $z_{\cal M}$ denote the representation of the original
and the compressed model ${\cal M}$, respectively;
$p(z_o)$ and $p(z_{\cal M})$ are the distributions of $z_o$ and $z_{\cal M}$, respectively.
A small $D_{\rm KL}$ indicates a closer relationship between
$p(z_{\cal M})$, the representation for the compressed model and $p(z_o)$,
the representation of the original model.

\subsection{Estimate Method of Kernel Width Parameter of Hidden Neuron }

 To estimate the mutual information between $Z_k$ and $Z_\ell$, we compute
\begin{equation}\label{eq:renyi}
I(Z_k; Z_\ell) = S_{\alpha}(A) +S_{\alpha}(B) - S_{\alpha}(A, B), 
\end{equation}
where $A$ and $B$ are Gram matrices of $Z_k$ and $Z_\ell$, respectively.
$S_{\alpha}(A)$ and $S_{\alpha}(B)$ denote the matrix-based 
R$\acute{e}$nyi’s $\alpha$-order \textit{entropy}, respectively.
$S_{\alpha}(A, B)$ denotes the \textit{joint entropy} of $Z_k$ and $Z_\ell$.

The estimation is analogous to the estimation of the mutual information
in a reproducing kernel Hilbert space (RKHS), that is,
solving the eigenvalues of the kernel matrix
to estimate the entropy, and solve the eigenvalues of
the Hadamard (entry-wise) product
of two kernel matrices to estimate the joint entropy. 
In this computation,
one of the key steps is to choose an appropriate kernel width parameter $\sigma$
to have a decent estimation.

To have it focus on a single neuron-related calculation,
we have to adjust the estimation by integrating
the optimal criterion and Scott's rule to estimate the kernel width parameter
if one hidden neuron is focused. 
The process is illustrated as follows.

First, We apply Scott's rule to estimate the kernel width parameter $\sigma^\ell$
of the focused hidden layer, we have
%
\begin{equation}\label{eq:sigma_l}
\sigma^\ell = \gamma N^{\frac{-1}{(4+d)}} \, , 
\end{equation}
where $N$ denotes the number of samples, $d$ denotes the the number of hidden neurons,
and $\gamma$ is an empirically determined constant.
%
%
Second, we adopt Eq.~\ref{eq:gaussian}
to calculate the kernel matrix $K_{\sigma^\ell}$ of hidden layer.
The RBF kernel is written as:
\begin{equation}\label{eq:gaussian}
K_\sigma ({\bf x}_i,{\bf x}_j) = \exp\left(-\frac{||{\bf x}_i -
{\bf x}_j||_{F}^{2} }{2 {\sigma}^2}\right),
\end{equation}
where ${{\bf x}_i,{\bf x}_j}$ denote data points, $ \left\| \cdot \right\|_{F}$ denotes the Frobenius norm, $\sigma$ is the kernel width parameter.
%
%
Third, we align kernel matrix $K_{\sigma^\ell}$ with kernel matrix $K_{\sigma^n}$ 
(kernel matrix of hidden layer) by maximizing the kernel alignment loss between these two kernel matrices, the kernel alignment loss~\cite{cristianini2001kernel} is written as: 

%
\begin{equation}
\label{eq:alignment}
  A(K_{\sigma^\ell}, K_{\sigma^n})=\frac{\left< K_{\sigma^\ell}, K_{\sigma^n}\right>_{F}}
  {\left\| K_{\sigma^\ell}\right\|_F \left\| K_{\sigma^n}\right\|_F }  \, , 
\end{equation}
where $ \left\| \cdot \right\|_{F}$ and
$\left< \cdot, \cdot\right>_F$
denote the Frobenius norm and inner product, respectively.

Thus, we choose our optimal $\sigma^n$ as Eq.~\ref{eq:sigma_n}:
\begin{equation}
\label{eq:sigma_n}
    \sigma^{n*} = \mathop{\arg\max}\limits_{\sigma{^n}}  A(K_{\sigma^l}, K_{\sigma^n})\, .
\end{equation}
The best choice of $\sigma^n$ according to different size
of mini-batches, we consider the approach from
Wickstr{\o}m et al.~\cite{wickstrom2019information}
to compute the exponential moving average to decide the final value,
such as
%
%
\begin{equation}\label{eq:sta_alpha}
\sigma_{n,t} = \beta \sigma_{n,t-1} + (1- \beta ) \sigma_{n, t}^{*}\, ,
\end{equation}
where $\beta \in [0, 1]$ and $\sigma_{n,1} = \sigma_{n, 1}^{*}$.
Finally, we obtain the kernel width parameter of every hidden neuron in the hidden layer.

\section{Experiment Result}
\label{sec:exp}

We conduct a few series of experiments to evaluate the effectiveness of the proposed model.
The first goal is to understand how accurate the mutual information estimation
on the focused hidden neurons is. After that, we need to confirm the effectiveness
of the proposed pruning method.
%
%
Relative FLOPs are used to indicate the compression level of the model,
the formula is expressed as:
$$\mbox{Relative FLOPs} = \frac{\mbox{FLOPs of pruned model}}{\mbox{FLOPs of original model}} \, .$$
The smaller the value of Relative FLOPs, the higher the compression level of the model. We follow the widely adopted definition of Relative FLOPs~\cite{khaki2024the, kwon2022fast}.

\subsection{Experimental Settings}

We evaluate the effectiveness of the proposed methods using the BERT-tiny model~\cite{turc1908well} on
the General Language Understanding Evaluation (GLUE)~\cite{wang2018glue} benchmark.
The BERT-tiny is a (pretraining $+$ fine-tuning) model from~\cite{turc1908well}, 
which consists of one Embedding layer and two transformer encoder blocks, with a hidden size of 512 for the FC layer in FFN. 
The GLUE benchmark contains a collection of NLU tasks and we fine-tuned it
on five downstream tasks: Single-Sentence Task (SST-2~\cite{socher2013recursive}),
Similarity and Paraphrase Tasks (STS-B~\cite{cer2017semeval}, MRPC~\cite{dolan2005automatically},
and QQP~\cite{dolan2005automatically}), and Inference Task (QNLI~\cite{dolan2005automatically}).
The batch size was set to 8, 8, 8, 16, and 16 for these tasks, respectively.
Additionally, the learning rate was set to $5\cdot10^{-4}$, $5\cdot10^{-4}$, $3\cdot10^{-4}$,
$3\cdot10^{-4}$, and $3\cdot10^{-4}$ for these tasks, respectively. Throughout all experiments,
we trained the model using the AdamW optimizer~\cite{loshchilov2018decoupled} with $\beta_1=0.9$,
$\beta_2= 0.999$, $\epsilon=10^{-8}$, and conducted a total of 4 fine-tuning epochs.
The overall data statistics and corresponding evaluation metrics are shown in Table.~\ref{tab:dataset}.
\begin{table}[tb]
  \begin{center}
  \footnotesize
  \setlength{\abovecaptionskip}{0cm}
   \caption{Data statistics of GLUE datasets. Among them, except for STS-B, which is a regression task,
   the others are classification tasks.}
  \resizebox{\linewidth}{!}{  
    \begin{tabular}{|l|c|c|c|c|c|}
    \hline
      Tasks          & Datasets     &  Training     & Validation     & Test & Metrics   \\
    \hline
    Single-sentence  & SST-2 &  67350 & 873 & 1821  & Accuracy\\ 
    \hline
    Inference  & QNLI & 104743 & 5463 & 5461 & Accuracy\\
    \hline
    \multirow{3}*{ {\makecell[l]{Similarity and\\ paraphrase}}}  & STS-B  & 5749 & 1379 & 1377 & {\makecell[c]{Pearson correlation ($r$)\\ Spearmen correlation ($r_s$)}}   \\                     
    \cline{2-6}
                            & MRPC  & 3668 & 408 & 1725 & F1 / Accuracy\\
                       \cline{2-6}
                 & QQP & 363870  & 40431 & 390965   & F1 / Accuracy\\         
     \hline
  \end{tabular}
  }
    \label{tab:dataset}
  \end{center}
\end{table}

In the estimation of mutual information settings,
we random sample 1$\%$  of the number of samples per task
training dataset to calculate the mutual information,
so we can know that the $N$ in Eq.~\ref{eq:sigma_l} 
is equal to the random sample number.
At the same time, in Eq.~\ref{eq:renyi}, we set $\alpha$ = 1.01,
in Eq.~\ref{eq:sigma_l}, we set $\gamma$ =1, $n$ =512.
The batch size of the calculation process is set 
to 100, which is the same as ~\cite{sarvani2022hrel}. Due to the randomness of MDS, we sample 500 random seeds and choose the best one based on Eq.~\ref{eq:kl}.

\subsection{The Results of Model Pruning}

In this part, we compare the proposed method to some
supervised learning approaches~\cite{yang-etal-2022-textpruner, kwon2022fast, park2023accurate},
self-supervised learning method~\cite{yang-etal-2022-textpruner}, 
and unsupervised learning methods ~\cite{li2016pruning, nova2023gradient},
overall three types of comparisons.
The proposed method is similar to the weight-magnitude approach~\cite{li2016pruning}
and KCM~\cite{nova2023gradient}, in the sense that all need no labeled data in the pruning procedure.
Moreover, the proposed method is a retraining-free approach,
which follows the convention from~\cite{kwon2022fast, nova2023gradient, park2023accurate}.
It is different from the approach adopted on~\cite{kwon2022fast, nova2023gradient, park2023accurate}
where weight-tuning on the left-out (unpruned) neurons is necessary
to confirm the more-than-acceptable network effectiveness.
Note that the random strategy has its output as an average of ten trials
to reveal the general behavior of the strategy.
In all experiments, we fix the pruned percentage to be $1\%$.

\begin{table}[tb]
  \begin{center}
  \footnotesize
  \setlength{\abovecaptionskip}{0cm}
   \caption{Results of different methods when the relative FLOPs equal to 40$\%$.  Abbreviations S, Self-S, and U denote the Supervised method, 
    the Self-supervised method, and the Unsupervised method, respectively.}
  \resizebox{\linewidth}{!}{
    \begin{threeparttable}
    \begin{tabular}{lrrrrrrr}
    \toprule
    Methods & {\makecell[c]{S/U\\/Self-S}}  &  {\makecell[c]{Relative\\FLOPs}}  & {\makecell[c]{SST-2\\Acc}} & {\makecell[c]{STS-B\\$r$/$r_s$}} & {\makecell[c]{MRPC\\Acc/F1}} & {\makecell[c]{QQP\\Acc/F1}} & {\makecell[c]{QNLI\\Acc}}  \\
    \midrule
    {\makecell[l]{BERT-tiny\\ (Original)}}  & & 100\% & 83.2 & 74.3/73.6 & 81.1/71.1  & 62.2/83.4 & 81.5  \\ 
    \midrule
    TextPruner~\cite{yang-etal-2022-textpruner} & S & 40\% & 80.8 & 72.9/70.5 & \textbf{81.3/70.7}  & 62.7/85.3 & \textbf{78.7}   \\
    Mask-tuning~\cite{kwon2022fast} & S & 40\% & 81.7 & 73.7/70.9 & 80.7/69.6 & 61.8/85.3 & 65.0 \\
    Kprune~\cite{park2023accurate} & S & 40\% &  \textbf{83.1} & \textbf{74.4/72.3}  & 81.0/70.1  &  61.8/84.2 & 77.5  \\
    \midrule
    TextPruner~\cite{yang-etal-2022-textpruner} & Self-S & 40\% & 81.8 & 70.3/68.7 & 80.8/70.0 & \textbf{62.8/84.9} & 76.2 \\
    \midrule
     {\makecell[l]{Random}}  & U& 40\% & 80.7 & 71.0/69.4 & 80.8/68.7 & 59.1/84.4 & 67.2 \\
     {\makecell[l]{Weight-Magnitude~\cite{li2016pruning}}} & U & 40\% & 81.8 & 71.4/69.6 & 80.8/68.5 & 61.2/83.9 & 67.7 \\
     {\makecell[l]{KCM~\cite{nova2023gradient}}} & U & 40\% & 78.8 & 72.6/70.3 & 81.1/69.8  & 61.9/84.0 & 74.5 \\
     {\makecell[l]{\textbf{Proposed}\\ \textbf{ method}}}  & U& 40\% &82.6  & 72.1/69.2 & 80.9/69.4 & 61.2/84.3 & 77.2 \\
     \bottomrule
  \end{tabular}
  
    \label{tab:comprasiom}
    \begin{tablenotes}
        \footnotesize
        \item[1] TextPruner~\cite{yang-etal-2022-textpruner}  includes two methods: one is self-supervised learning, and the other is supervised learning.
    \end{tablenotes}
   \end{threeparttable}
   }
  \end{center}
\end{table}

In Table.~\ref{tab:comprasiom}, we show that the proposed model
can maintain the performance of the original model
for various tasks given the relative FLOPs are set to be $40\%$.
Let us compare the proposed method to the Weight-Magnitude~\cite{li2016pruning}
and KCM~\cite{nova2023gradient} approaches,
both belonged to the category of unsupervised methods,
and own some similarities to the proposed method.
The proposed method performs superior to the Weight-Magnitude method
on the tasks SST-2 and the QNLI while slightly worse on the other three tasks.
On the other hand, we also compare the proposed method to
another self-supervised method like Yang et al.~\cite{yang-etal-2022-textpruner}.
The proposed method is superior to other methods on all but the MRPC and QQP tasks.
To compare to the category of supervised method,
we evaluate the superiority of the proposed method to other supervised methods.
In this category, the proposed method performs better than the other two methods
on the SST-2 and QNLI tasks. For instance, the proposed method is better than
Mask-tuning~\cite{kwon2022fast} on the SST-2 and QNLI tasks,
and better than Yang et al.~\cite{yang-etal-2022-textpruner} on the SST-2 task.
For other tasks, the proposed method and the two supervised methods perform similarly.

\begin{figure*}[tb]
\centering 
\includegraphics[width=1\textwidth]{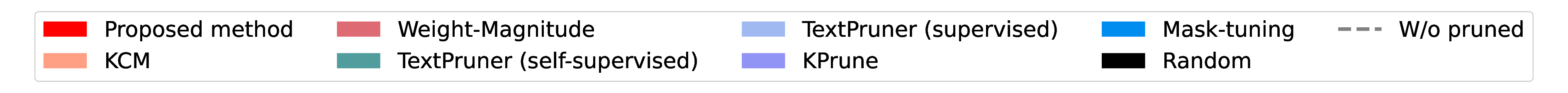}
\subfigure[SST-2]{     
\includegraphics[width=0.30\textwidth]{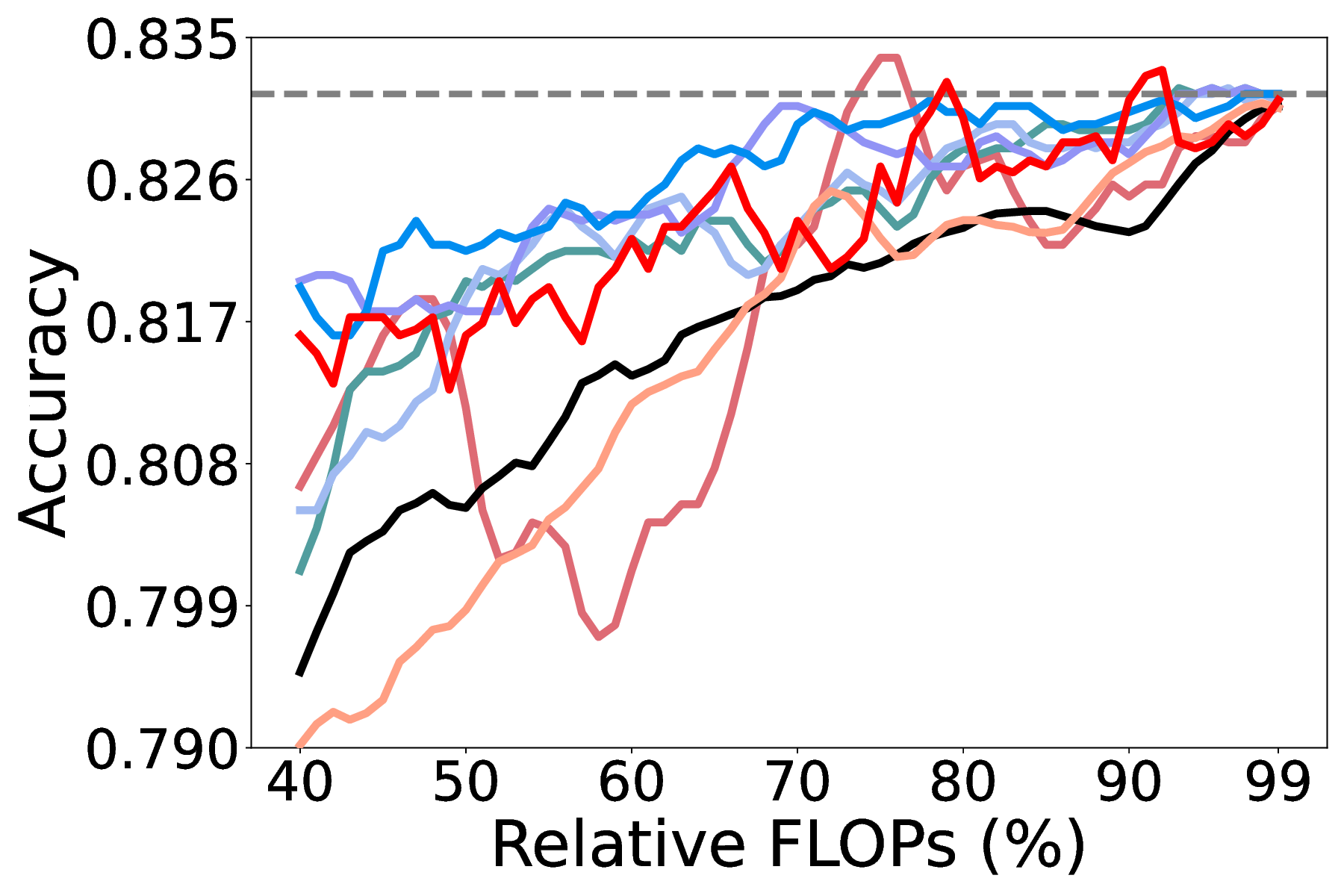}
}
\subfigure[QNLI]{     
\includegraphics[width=0.30\textwidth]{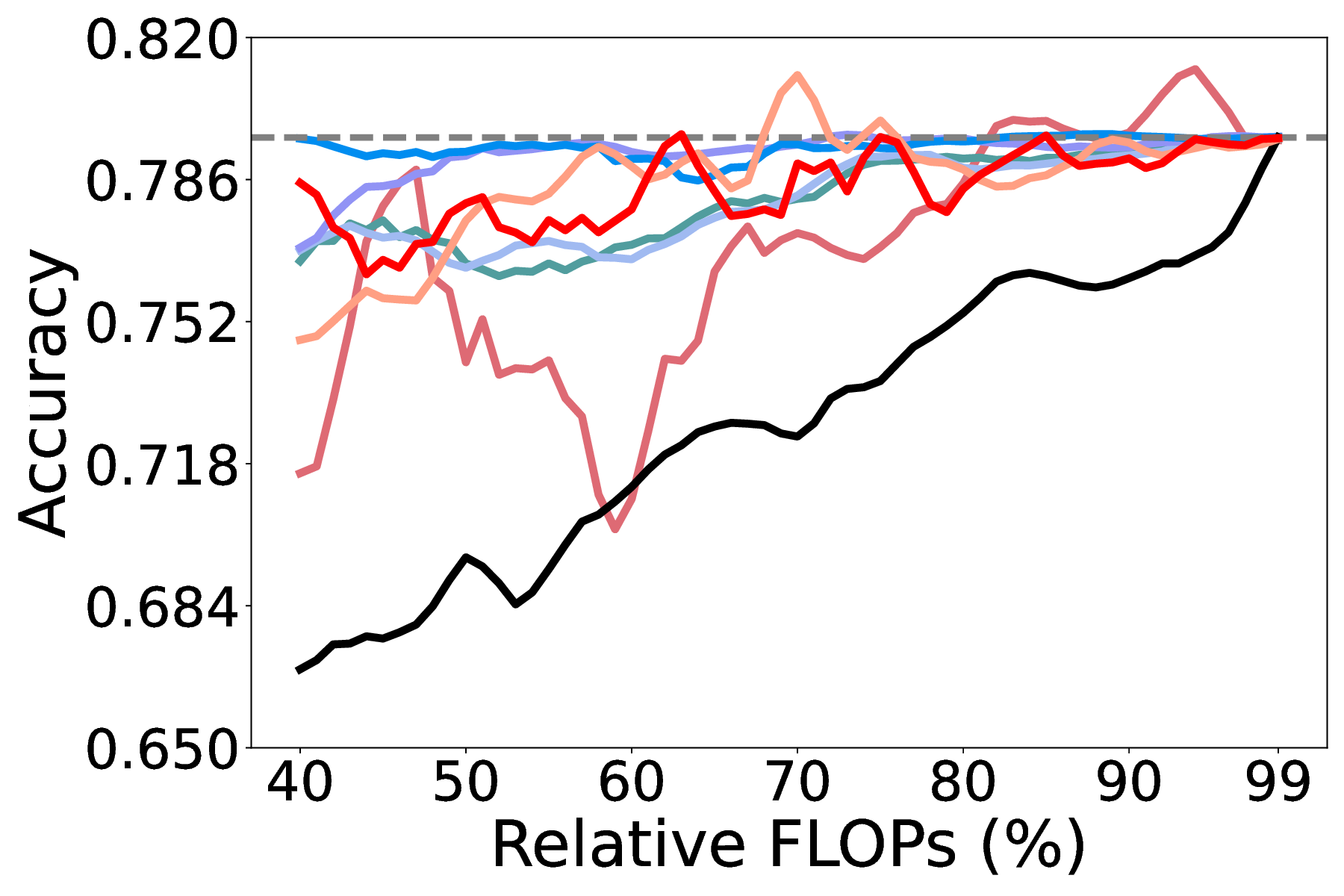}
}
\subfigure[STS-B]{     
\includegraphics[width=0.30\textwidth]{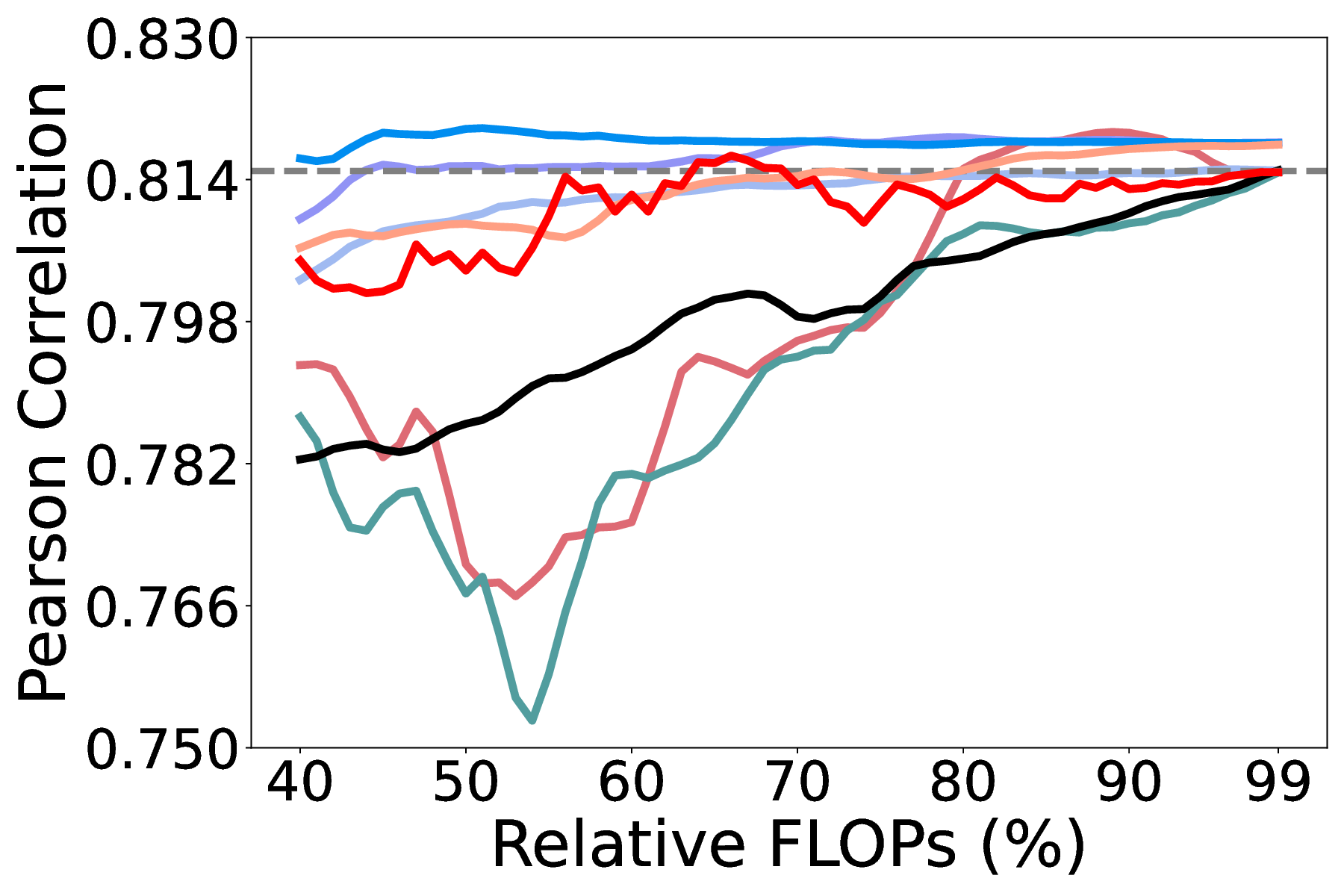}
}
\subfigure[MRPC]{     
\includegraphics[width=0.30\textwidth]{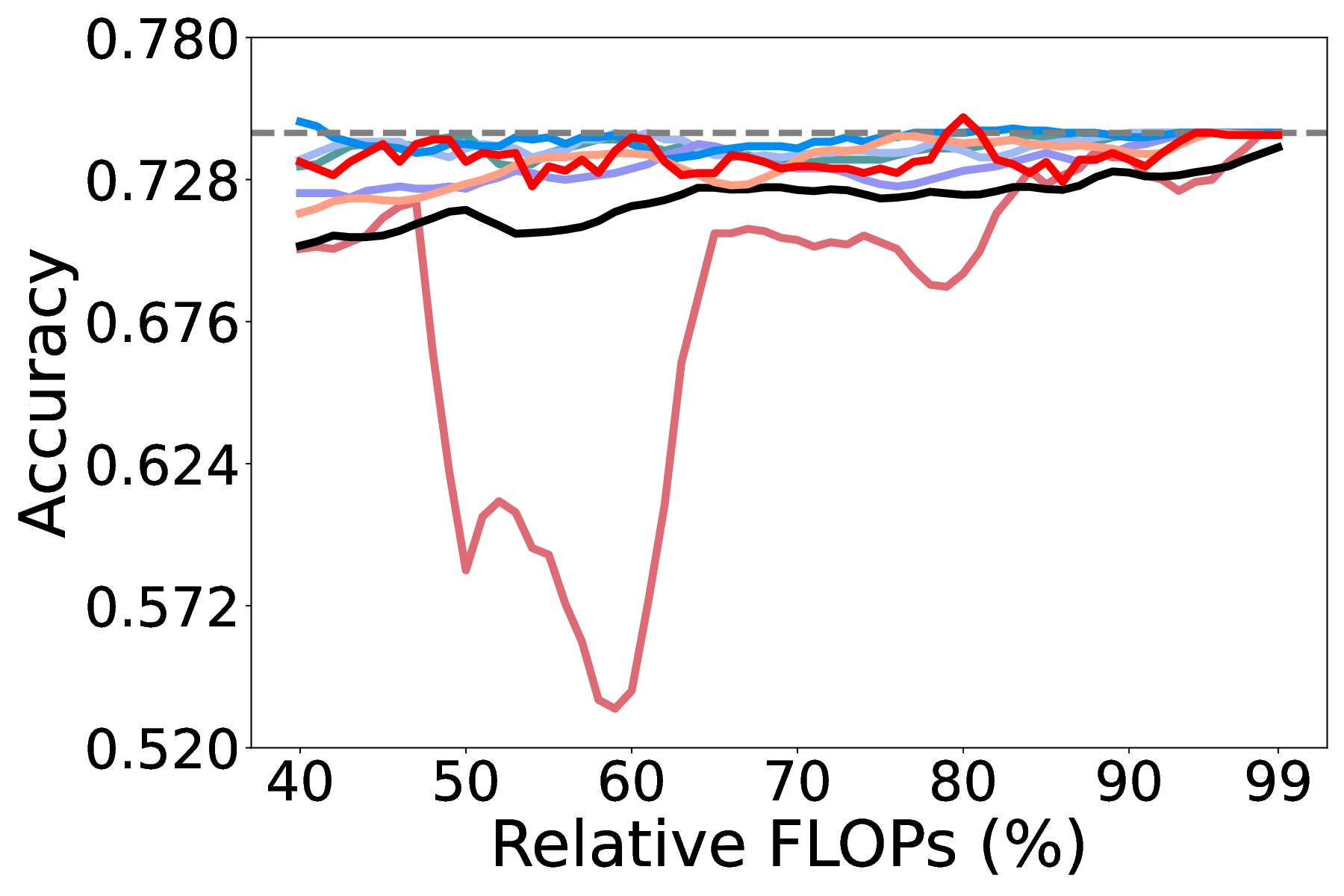}
}
\subfigure[QQP]{     
\includegraphics[width=0.30\textwidth]{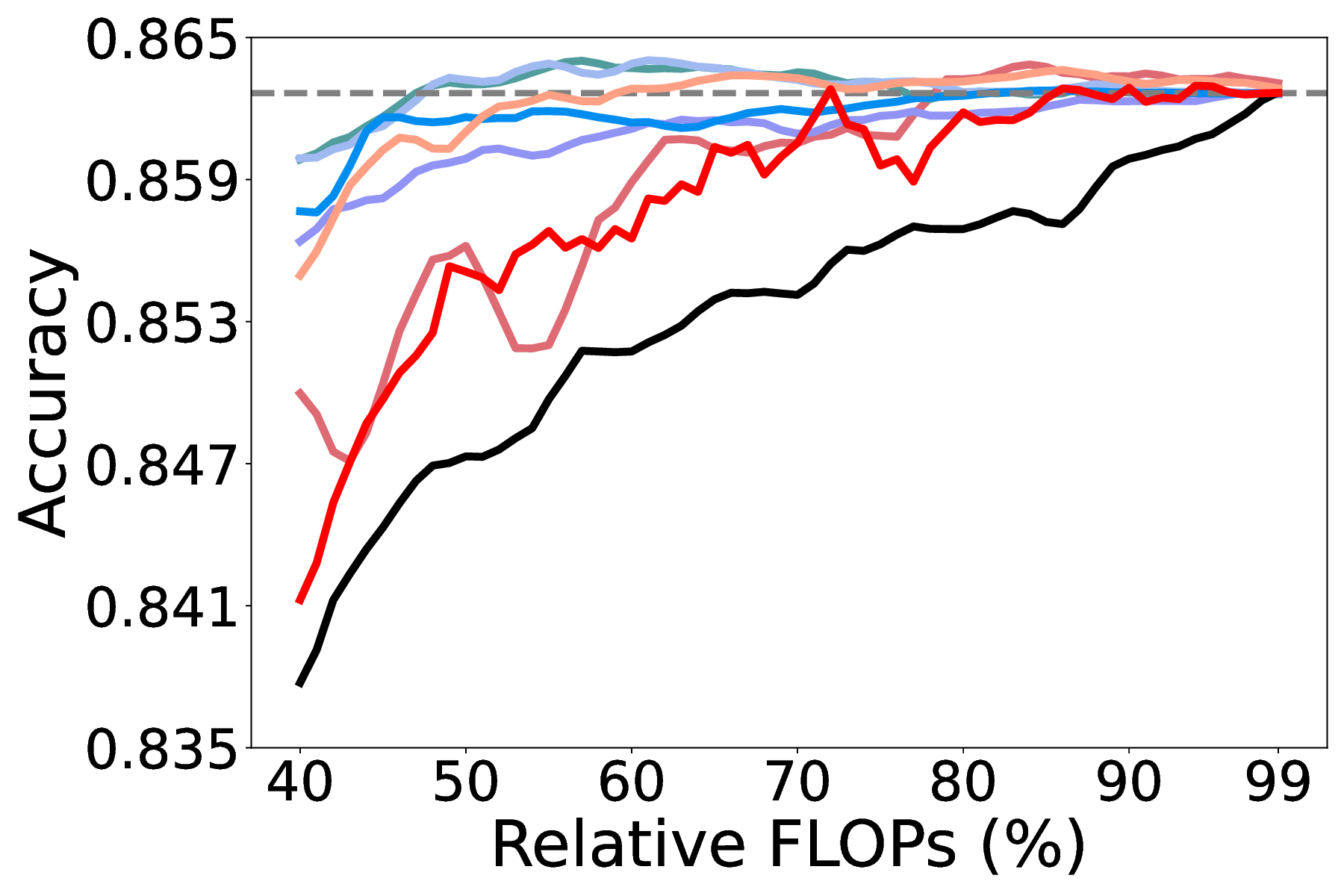}
}

\caption{Performance of the proposed method against other methods on BERT-tiny. Testing was conducted using the dev set to control relative FLOPs incrementally by percentage. }    
\label{fig:compari_diff_method}    
\end{figure*}

To clearly show the model performance under different compression rates,
we consider the following set of comparisons given five tasks.
As shown in Fig.~\ref{fig:compari_diff_method}, we demonstrate the result
for every $1\%$ change of the compression rate.
For all the tasks other than STS-B, the proposed method performs better than the random strategy.
If compared to the unsupervised learning method proposed by Li et al.~\cite{li2016pruning},
the proposed pruning method shows better performance on all except the QQP dataset,
and both perform similarly on the QQP dataset.
Compared to another method from the unsupervised group,
we consider KCM~\cite{nova2023gradient}. In this case, both perform similarly
on all four except the QQP dataset. The performance on SST-2 is slightly better
if by the proposed method.

When checking the self-supervised method such as Yang et al.~\cite{yang-etal-2022-textpruner},
the proposed method shows its advantage on both the QNLI and STS-B datasets,
but not on the SST-2 dataset.
In the category of supervised learning, the proposed method shows similar results
on both the STS-B and QQP datasets given low compression rates,
while the proposed method could perform poorly given the high compression rates.
On the SST-2, QNLI, and MRPC datasets, all behave similarly.

\subsection{Mutual Information between Hidden Neurons Estimation}

In this part of experiment, we take a close look about
the MI estimation between hidden neurons.
We compare the adopted approach and the estimation based on
the kernel width parameter tuning (Scott's rule).
Of course, no true value can be obtained to confirm
how accurate the MI estimation is, we provide the study on the model test accuracy
to provide another support for the pruning effectiveness.

\begin{figure*}[ht]
\centering 
\subfigure[SST-2]{     
\includegraphics[width=0.30\textwidth]{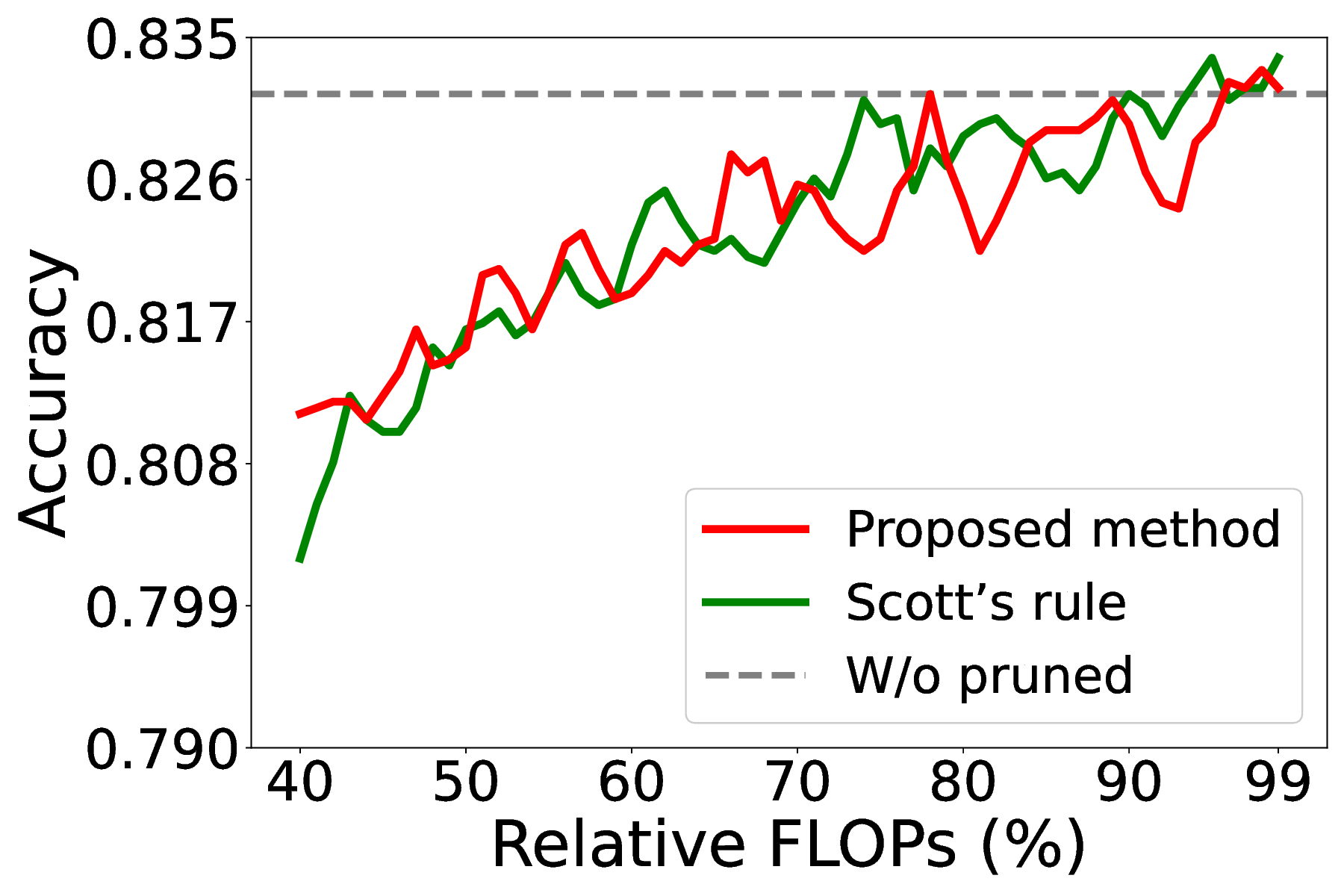}
}
\subfigure[QNLI]{     
\includegraphics[width=0.30\textwidth]{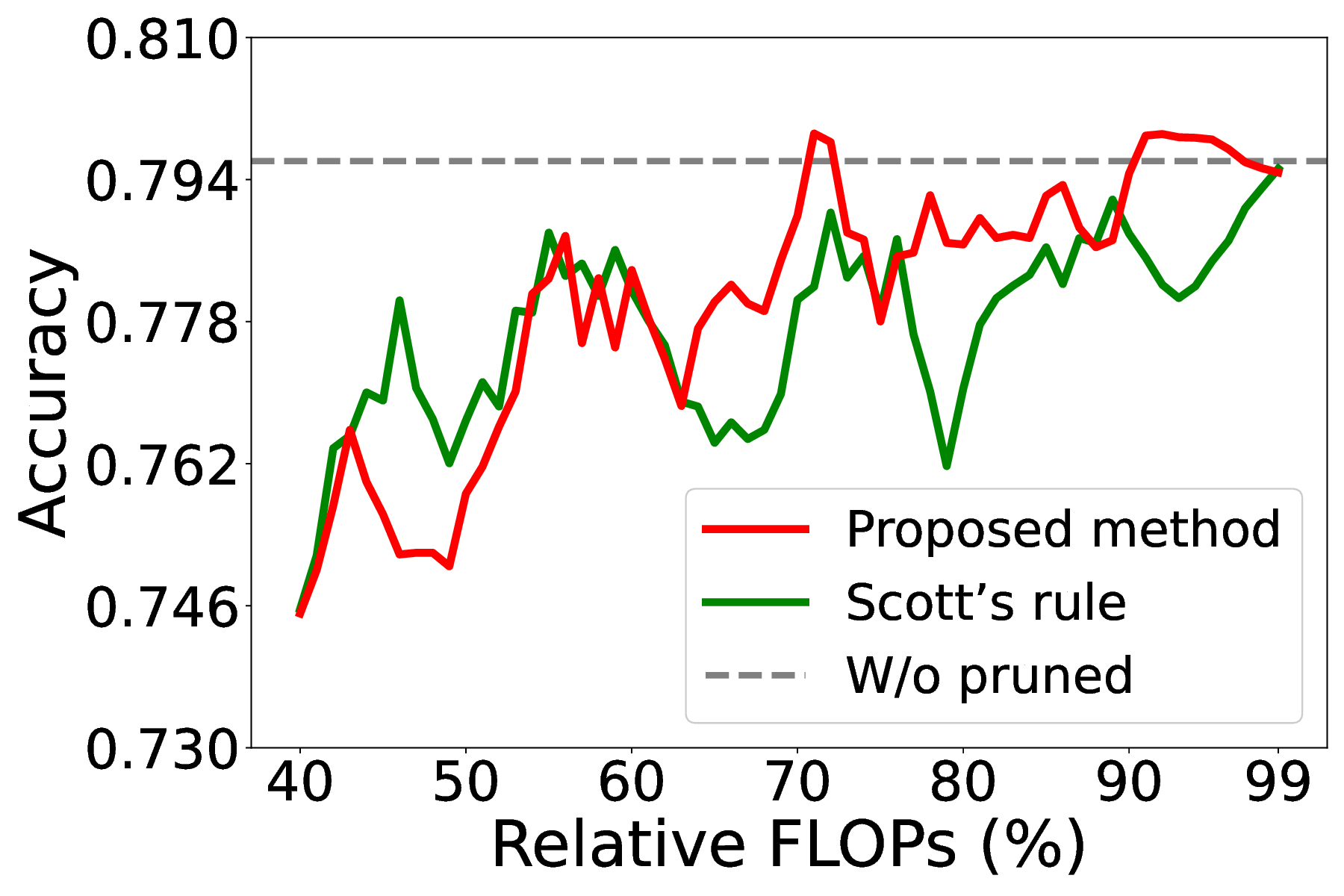}
}
\subfigure[STS-B]{     
\includegraphics[width=0.30\textwidth]{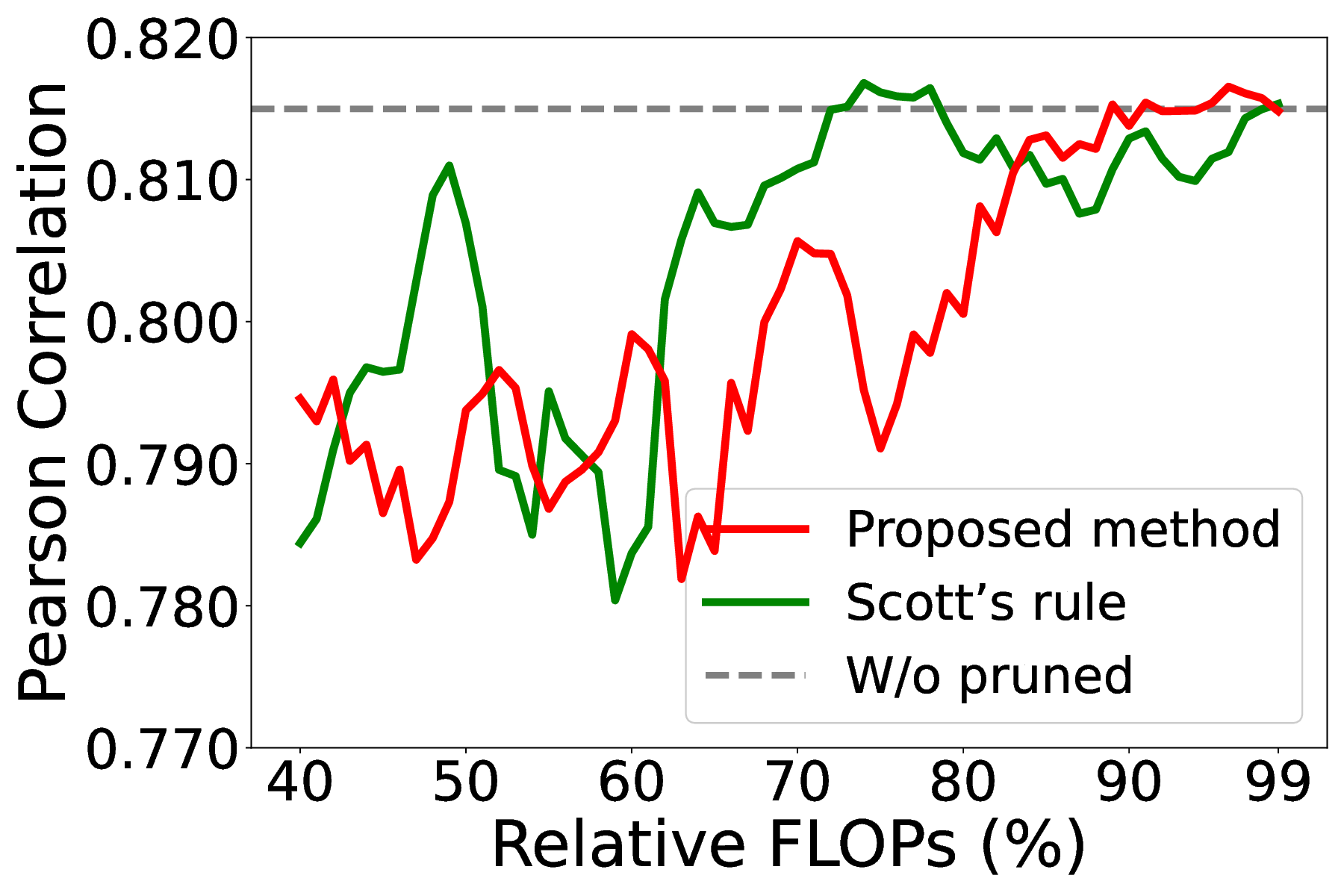}
}
\subfigure[MRPC]{     
\includegraphics[width=0.30\textwidth]{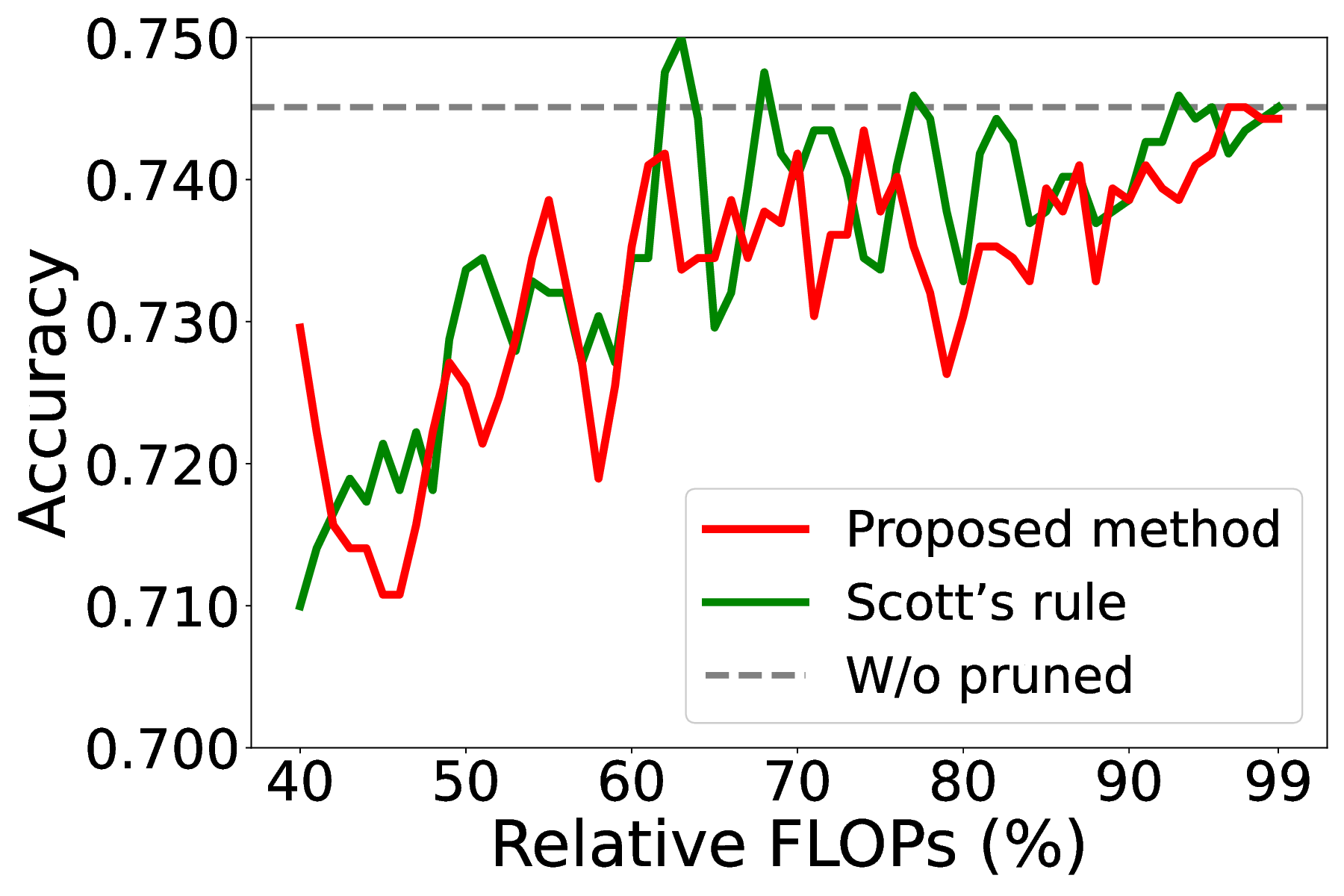}
}
\subfigure[QQP]{     
\includegraphics[width=0.30\textwidth]{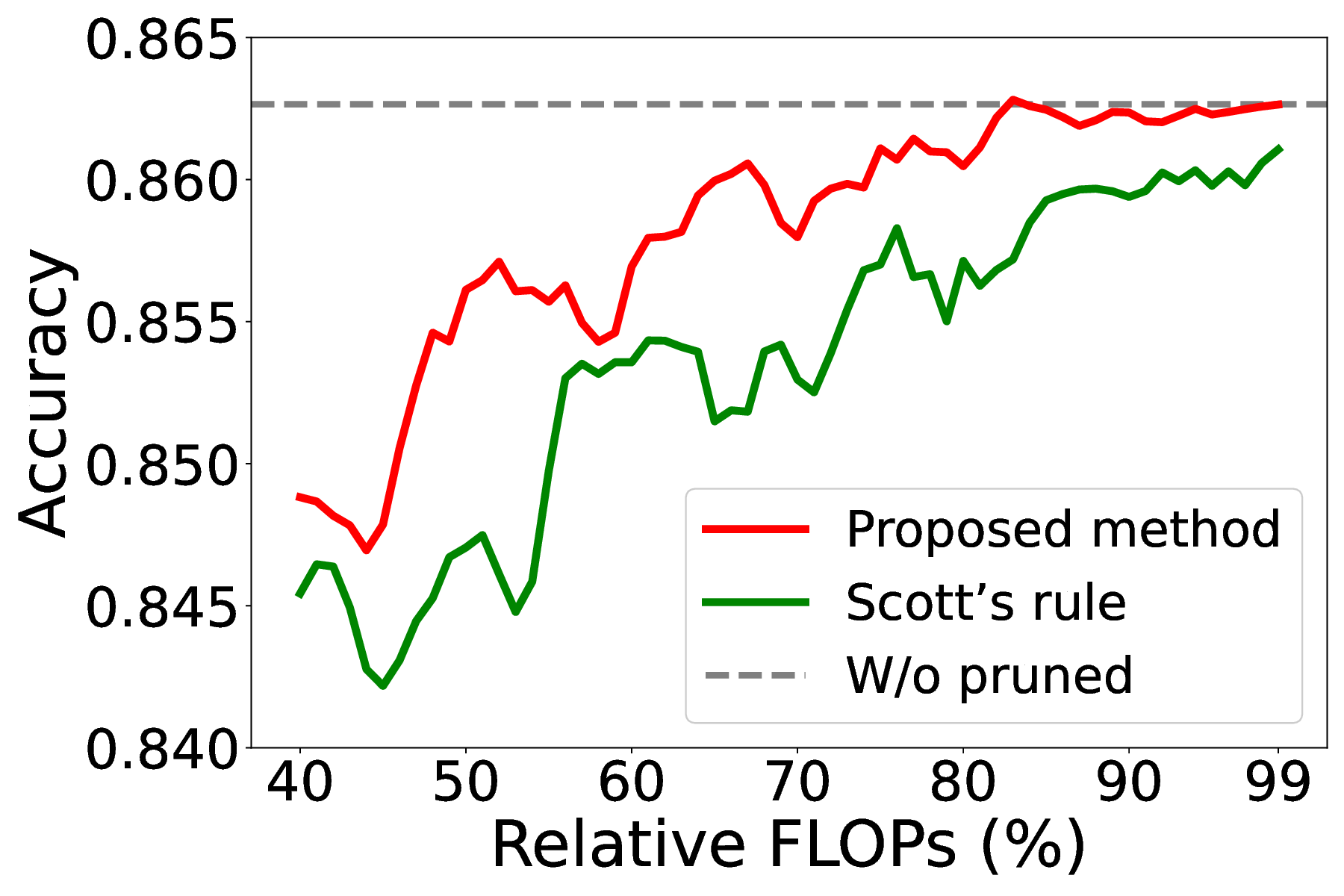}
}

\caption{Performance of the proposed method against other MI estimators on BERT-tiny. Testing was conducted using the dev set to control relative FLOPs incrementally by percentage. }    
\label{fig:mi_diff_method}    
\end{figure*}

As shown in FIG.~\ref{fig:mi_diff_method},
the proposed method offers better performance in terms of prediction accuracy
than other methods in most cases. We have a clear advantage
on the QQP dataset, across all compression rates. Working on the QNLI dataset,
the proposed method still keeps an advantage in most cases.
On the SST-2 and MRPC datasets, both
(the proposed method and Scott's rule) show similar results.
At the last, the proposed method may perform poorly on the STS-B dataset,
for at least the low compression rate cases.

\subsection{Ablation Study}

We take turns studying some alternative approaches
to see the possibilities of further improvement from the proposed method.

\subsubsection{Mutual Information vs. Pearson Correlation Coefficient}

\begin{figure}[tb]
\centering 
\subfigure[SST-2]{     
\includegraphics[width=0.30\textwidth]{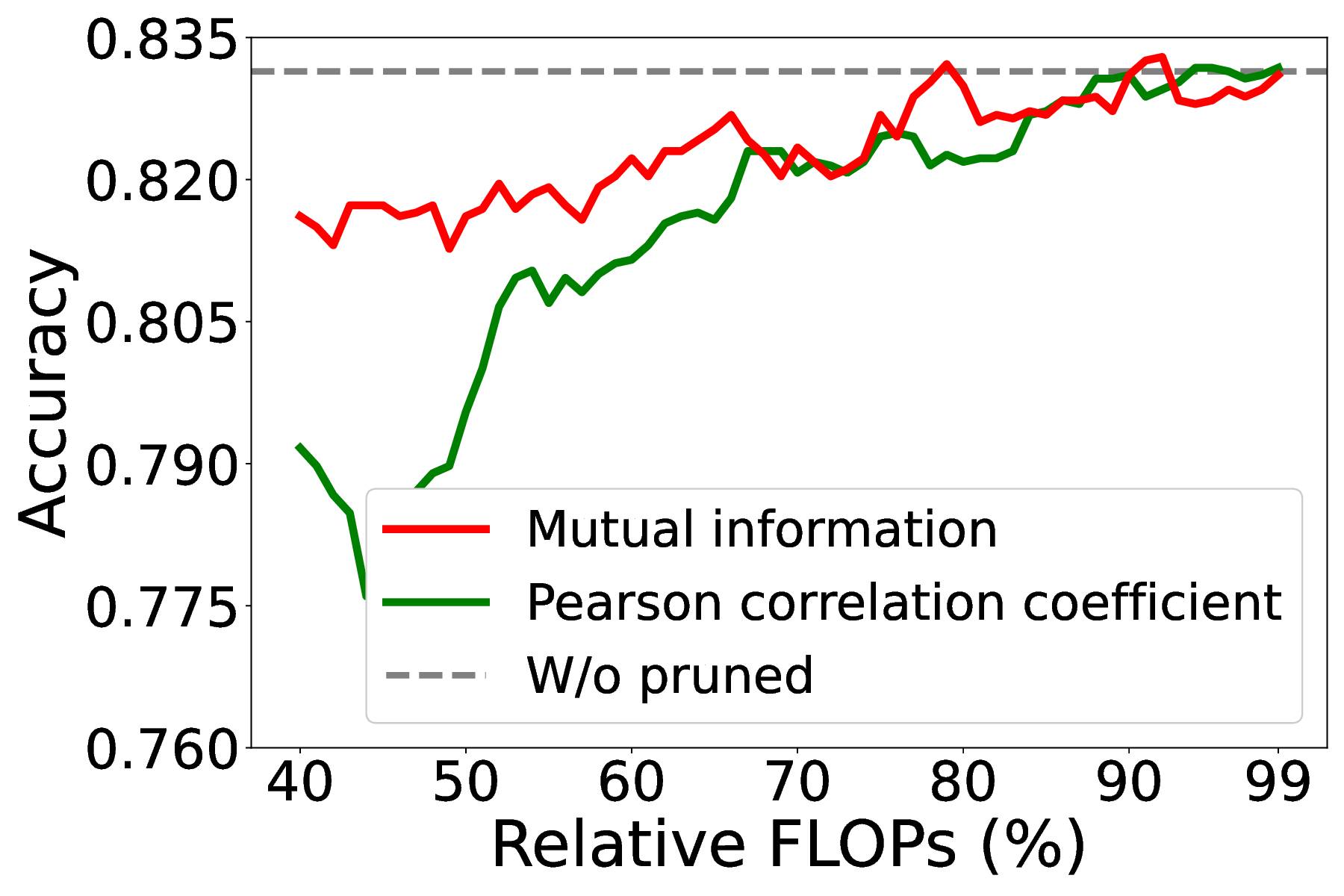}
}
\subfigure[QNLI]{     
\includegraphics[width=0.30\textwidth]{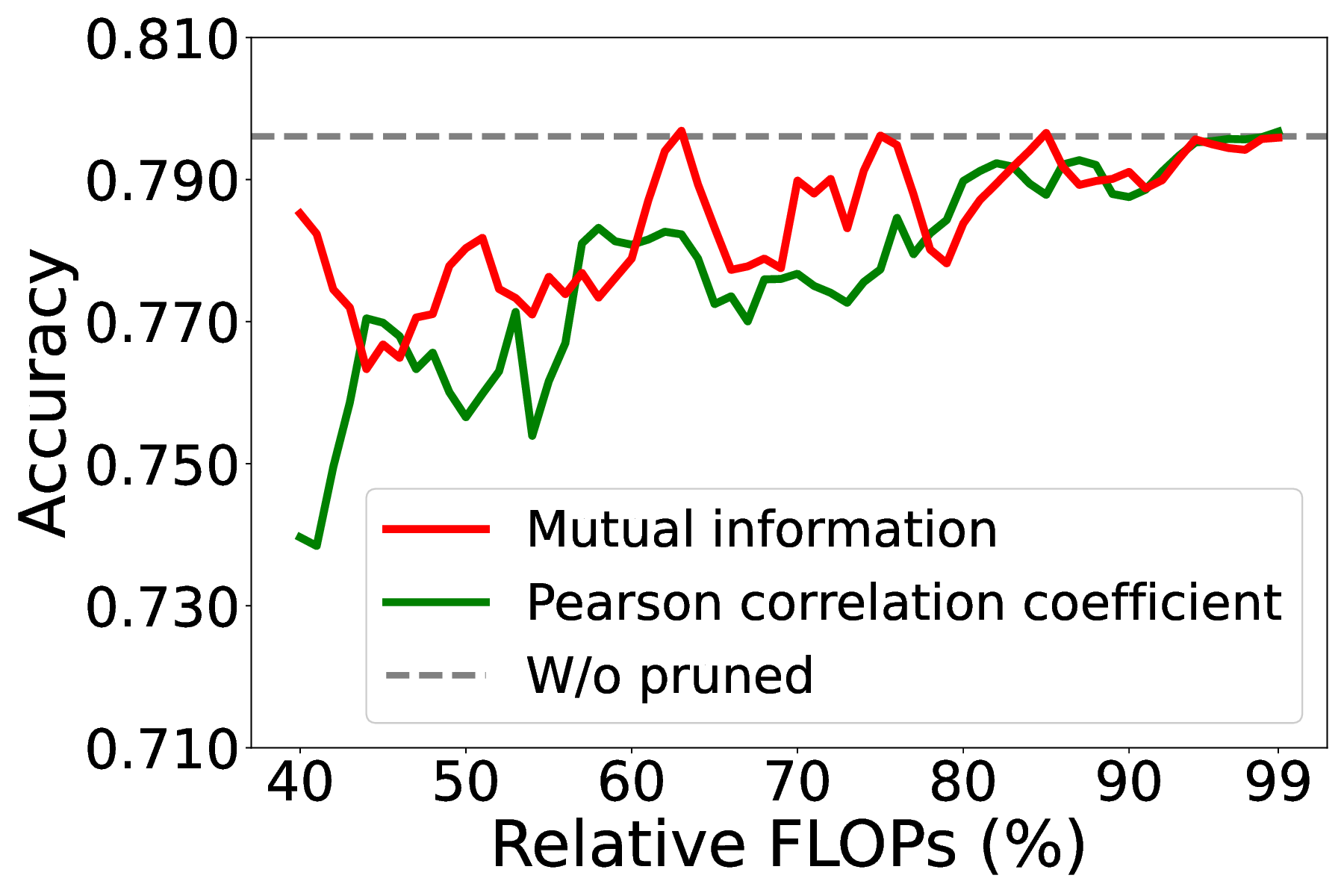}
}
\subfigure[STS-B]{     
\includegraphics[width=0.30\textwidth]{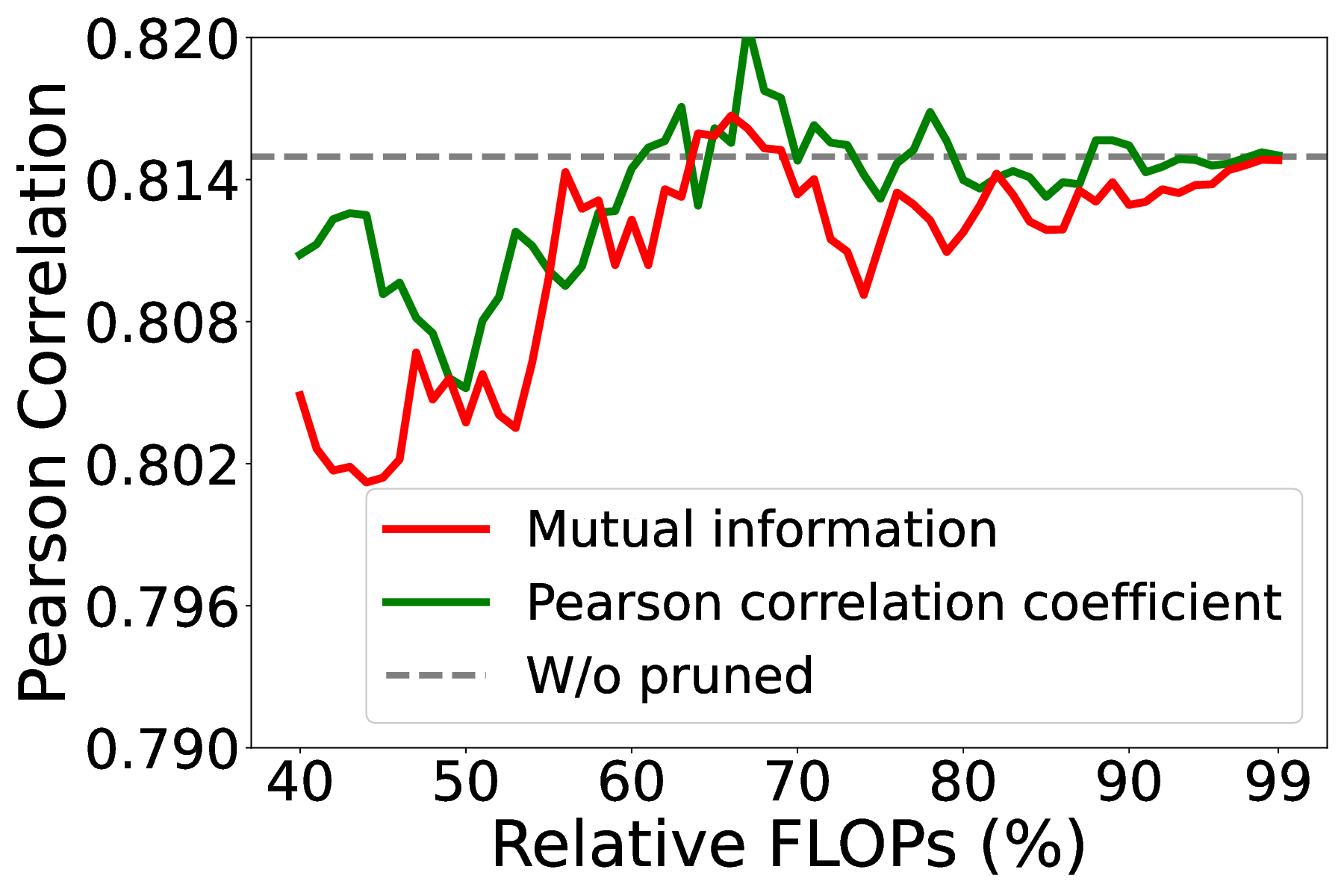}
}
\subfigure[MRPC]{     
\includegraphics[width=0.30\textwidth]{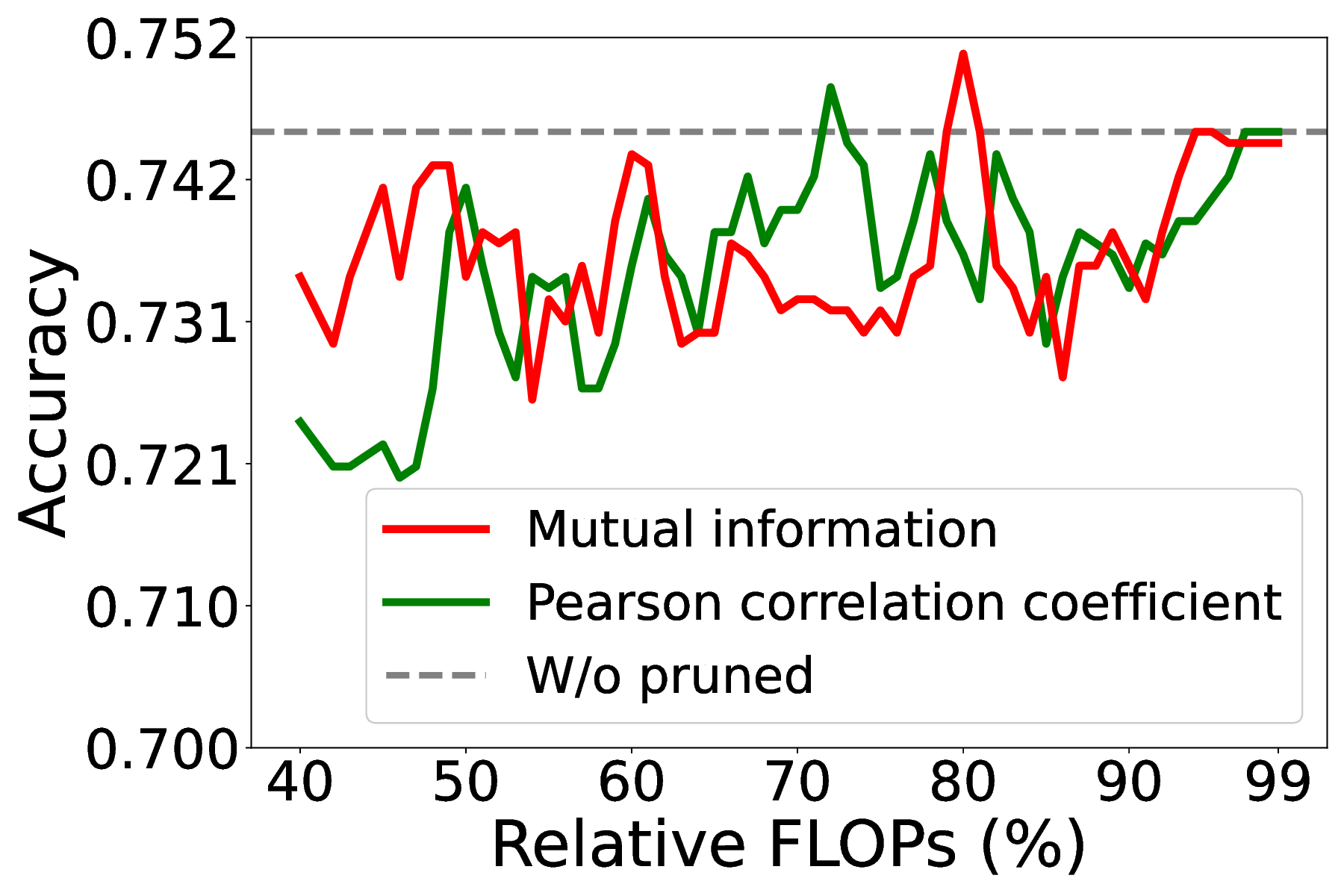}
}
\subfigure[QQP]{     
\includegraphics[width=0.30\textwidth]{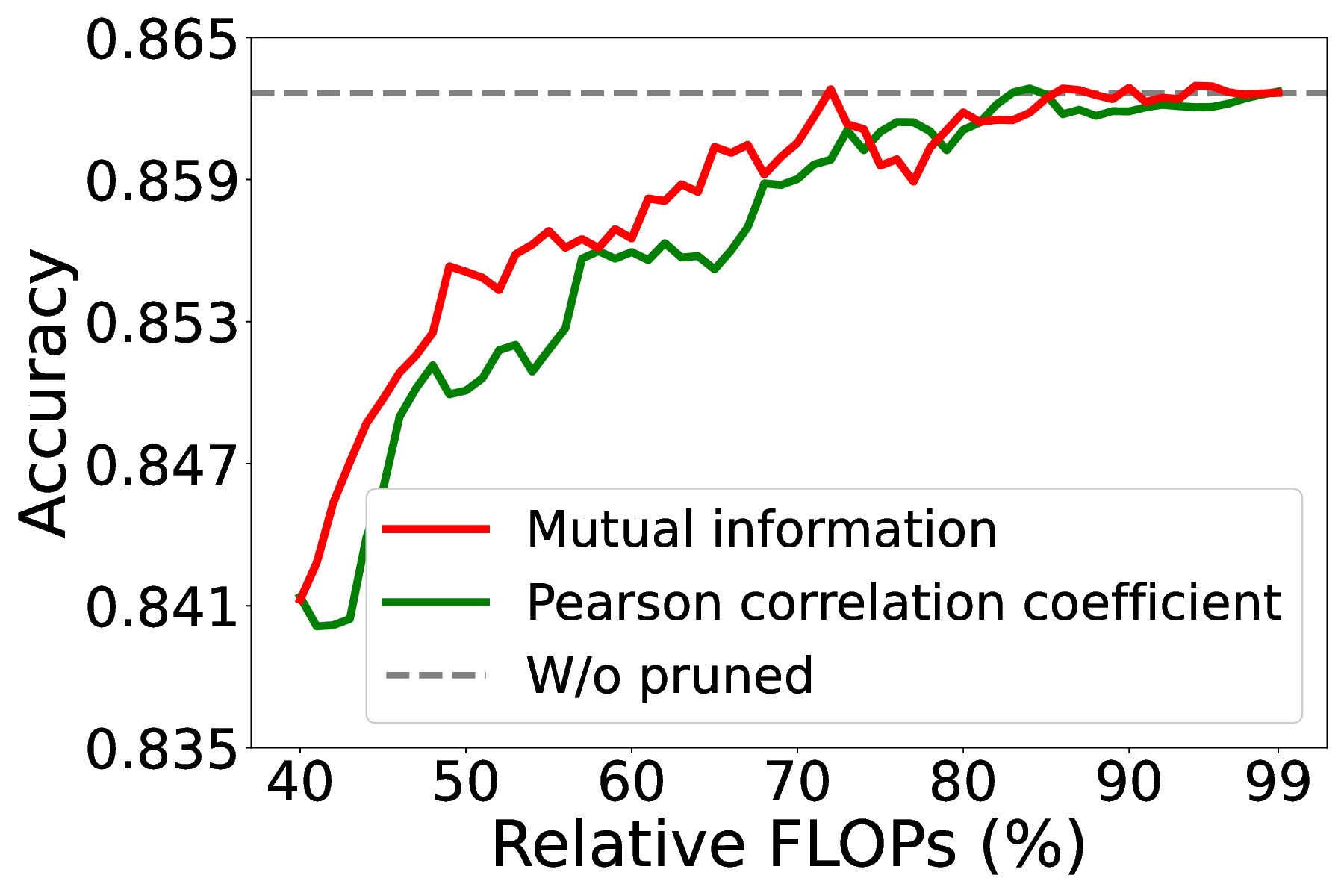}
}

\caption{The result based on mutual information or Pearson correlation coefficient computation}
\label{fig:mi_pcc}    
\end{figure}

First, we consider an alternative measure to describe how two variables
are related to each other, when we need to measure the relation
between two groups of neurons. For instance, we may choose
Pearson correlation coefficient (PCC) to substitute the role of MI
in Algorithm~\ref{alg:filter} to guide the pruning procedure.
The result from both, the MI-based and the PCC-based ones
are shown in FIG.~\ref{fig:mi_pcc}.

In FIG.~\ref{fig:mi_pcc}, using MI rather than PCC to describe
the relation between neurons performs the best
given the high-compression (small remaining model) situations,
for all but the STS-B dataset. It implies that MI can offer complex descriptions between different random variables,
compared to the one from PCC which covers only simple or linear relationships between random variables.
That is to say, the pruning on a large model may not need careful treatment because
neurons may find their substitutes easily, while the pruning on
a small model needs precise calculations to be done,
which could be offered from the MI-based approach.

\subsubsection{Data Samples for Mutual Information Estimation}

This series of experiments is devoted to studying how large data is enough
to estimate the MI value. If a small dataset offers a result similar
to the one from a large dataset, then we prefer using a small dataset to save
the running time in model pruning.

When estimating the MI value, we rely on the values of the hidden neurons.
That is, once the network is built after a converged training,
we can sample only a small portion of the input data, and use them to
activate the hidden neurons for the estimation. According to our study,
the estimation needs no large-scale data to confirm the pruning effect.

\begin{figure}[tb]
\centering 
\subfigure[SST-2]{     
\includegraphics[width=0.30\textwidth]{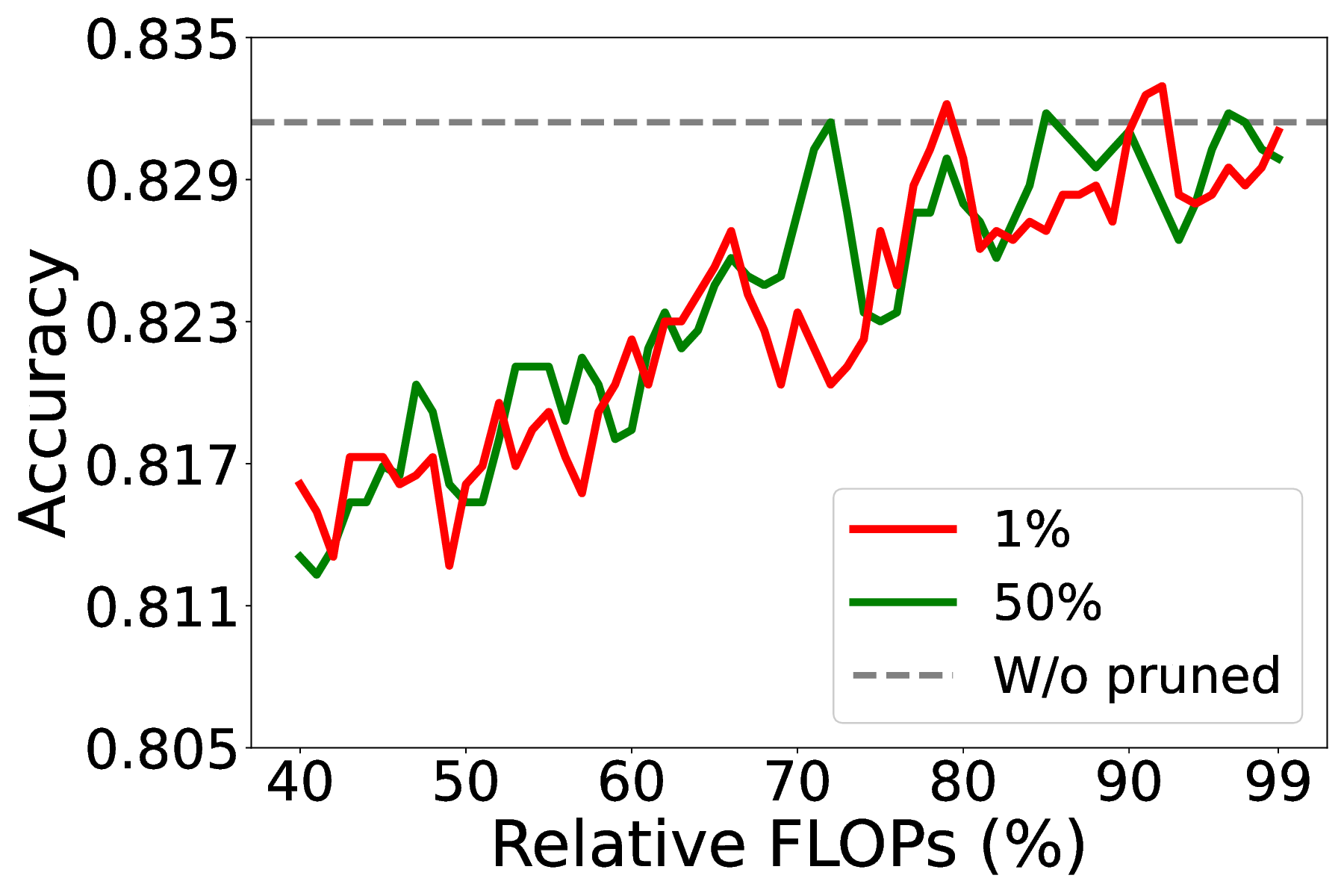}
}
\subfigure[QNLI]{     
\includegraphics[width=0.30\textwidth]{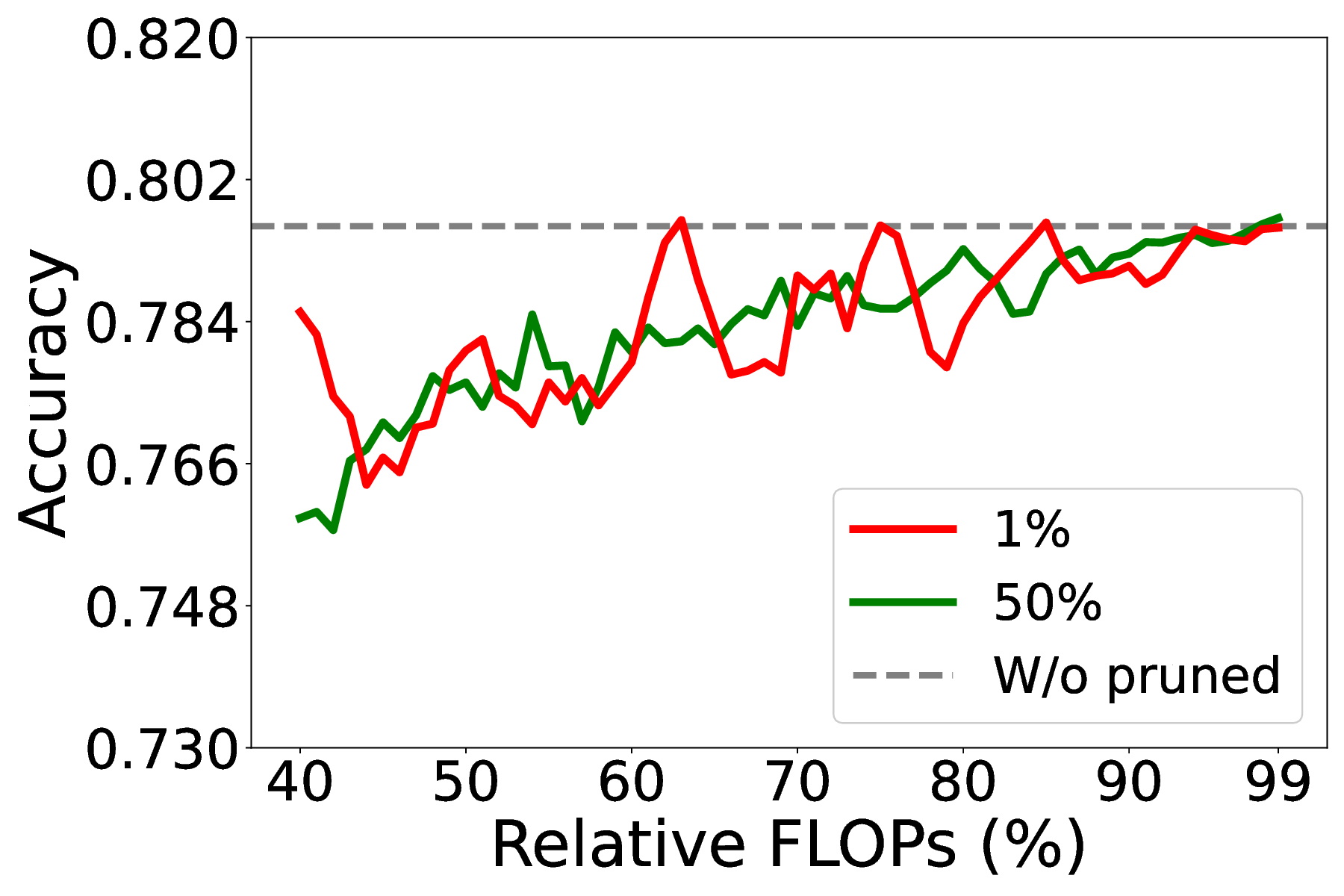}
}
\subfigure[STS-B]{     
\includegraphics[width=0.30\textwidth]{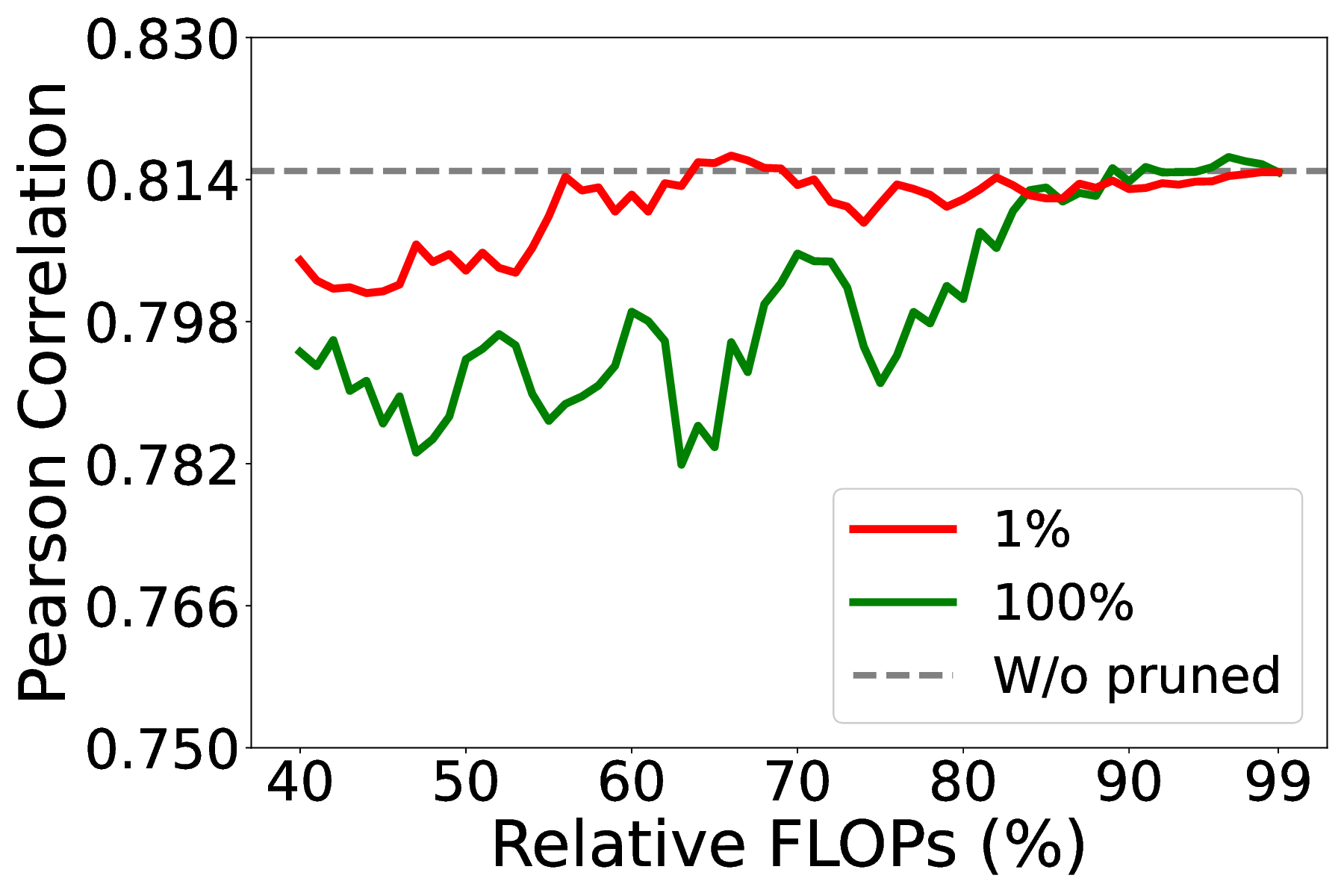}
}
\subfigure[MRPC]{     
\includegraphics[width=0.30\textwidth]{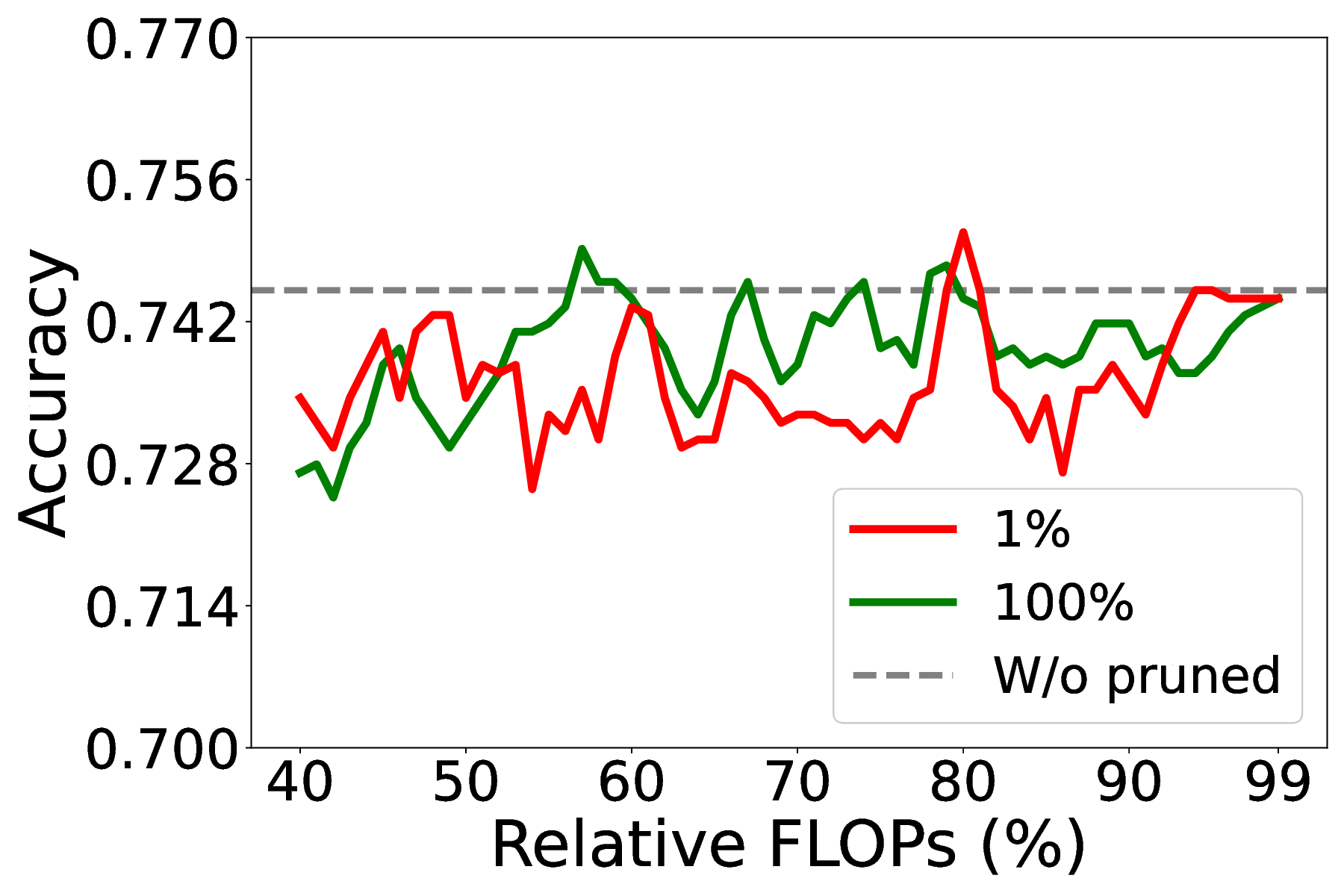}
}
\subfigure[QQP]{     
\includegraphics[width=0.30\textwidth]{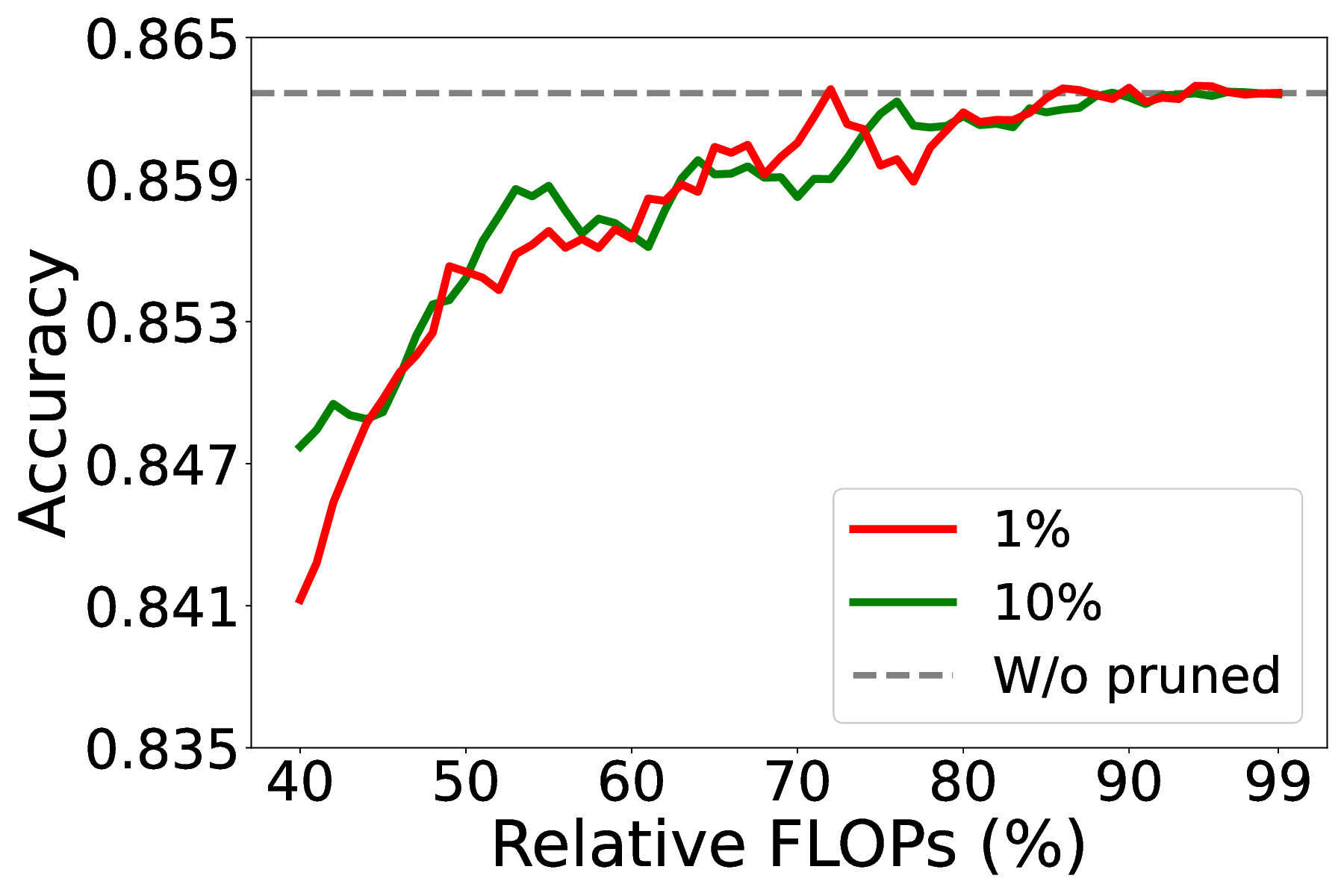}
}

\caption{Insignificant difference between the result from either the complete
or partial dataset.}
\label{fig:compare_on_data_samples}    
\end{figure}

As shown in Fig.~\ref{fig:compare_on_data_samples},
we take turns to consider $1\%$, $10\%$, $50\%$, or $100\%$ (all) of the complete data
to see their pruning performance. We feel lucky to see that
different data portions may still provide similar model performance from the pruned model,
on all but the STS-B dataset. Note that we need a regression task on the STS-B dataset,
while classification tasks are needed on all other datasets.

\subsubsection{Different \texorpdfstring{$\alpha$}{alpha} for Matrix-based R\'enyi's \texorpdfstring{$\alpha$}{alpha}-order Entropy Estimation}

\begin{figure}[tb]
\centering 
\subfigure[SST-2]{     
\includegraphics[width=0.30\textwidth]{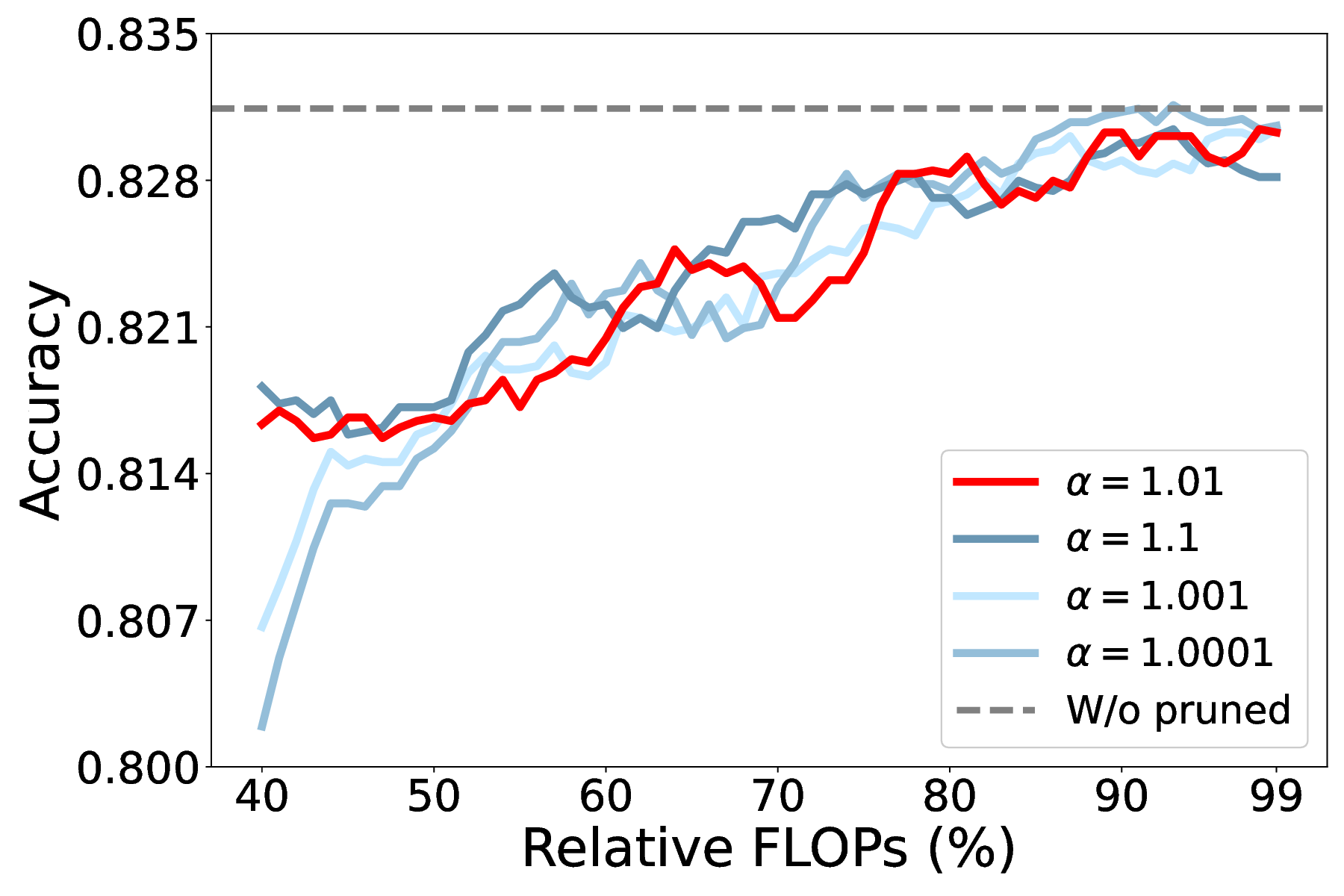}
}
\subfigure[QNLI]{     
\includegraphics[width=0.30\textwidth]{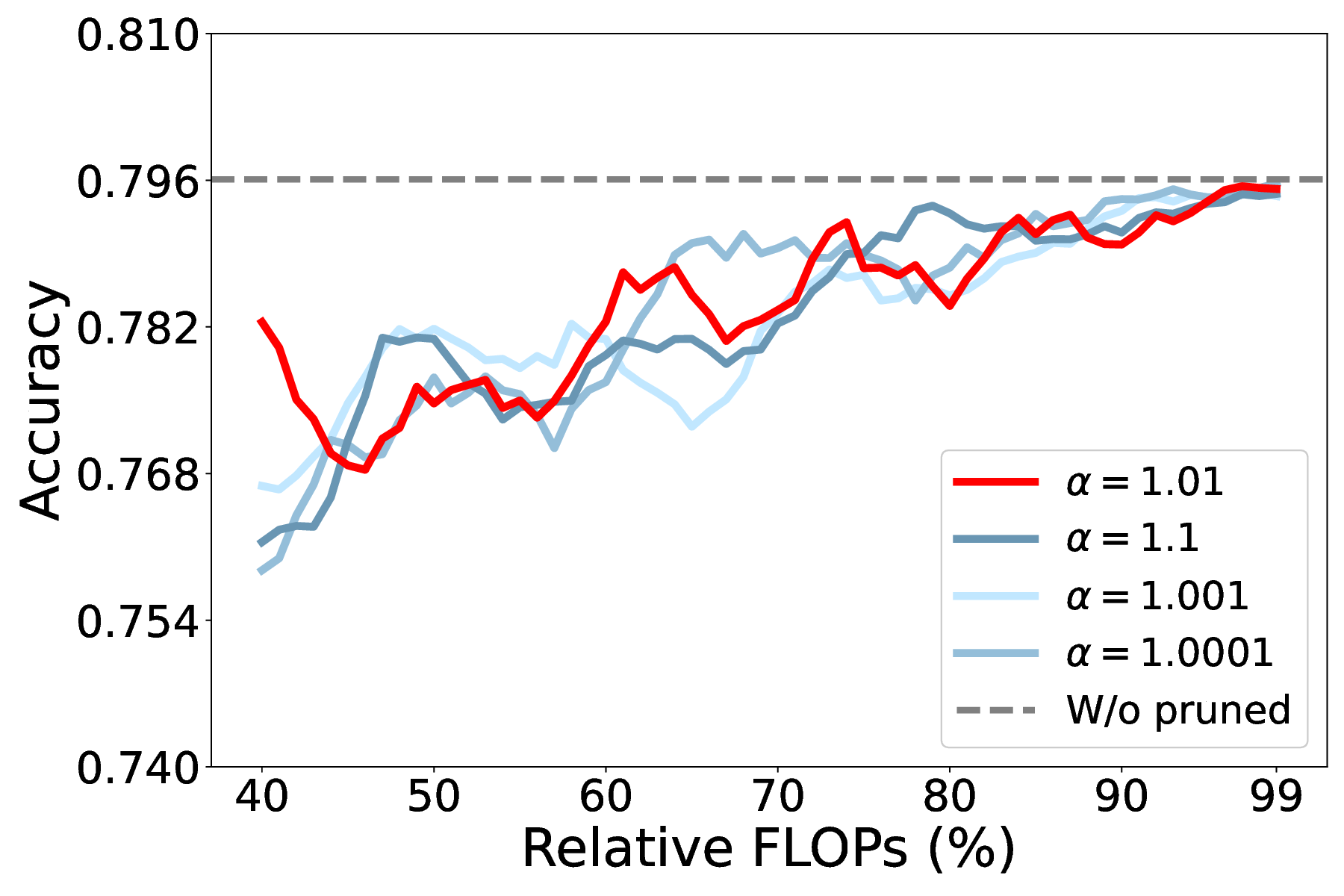}
}
\subfigure[STS-B]{     
\includegraphics[width=0.30\textwidth]{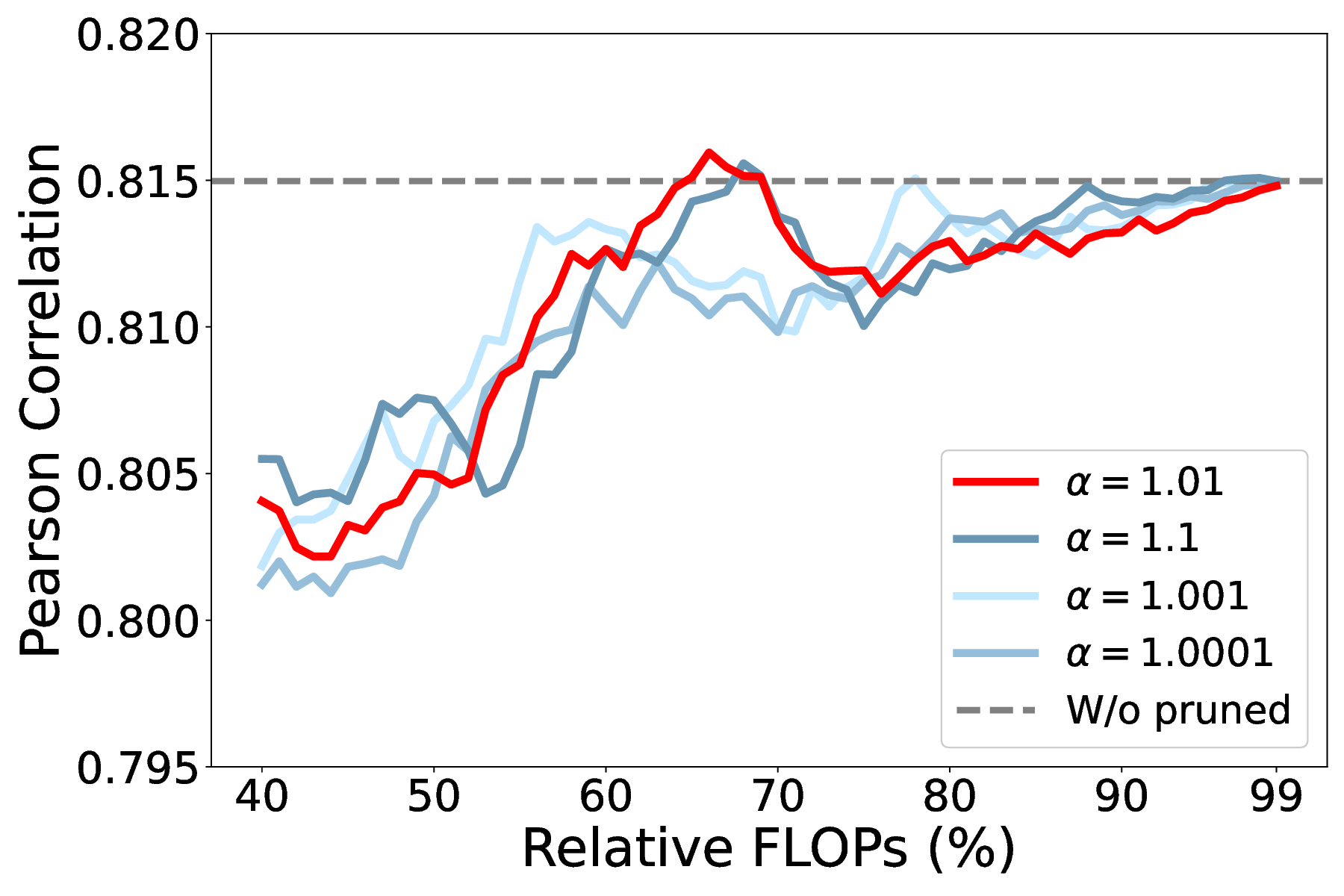}
}
\subfigure[MRPC]{     
\includegraphics[width=0.30\textwidth]{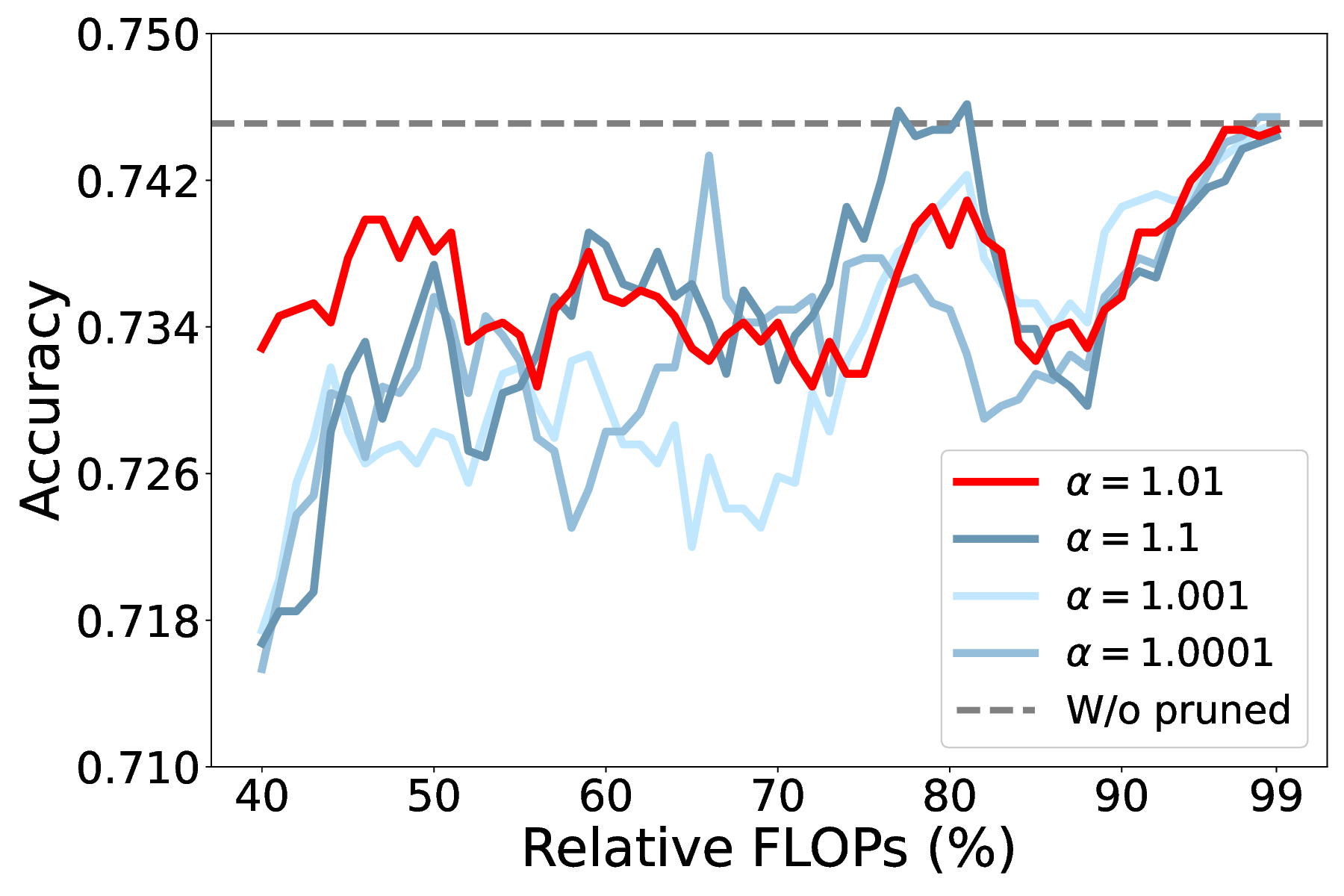}
}
\subfigure[QQP]{     
\includegraphics[width=0.30\textwidth]{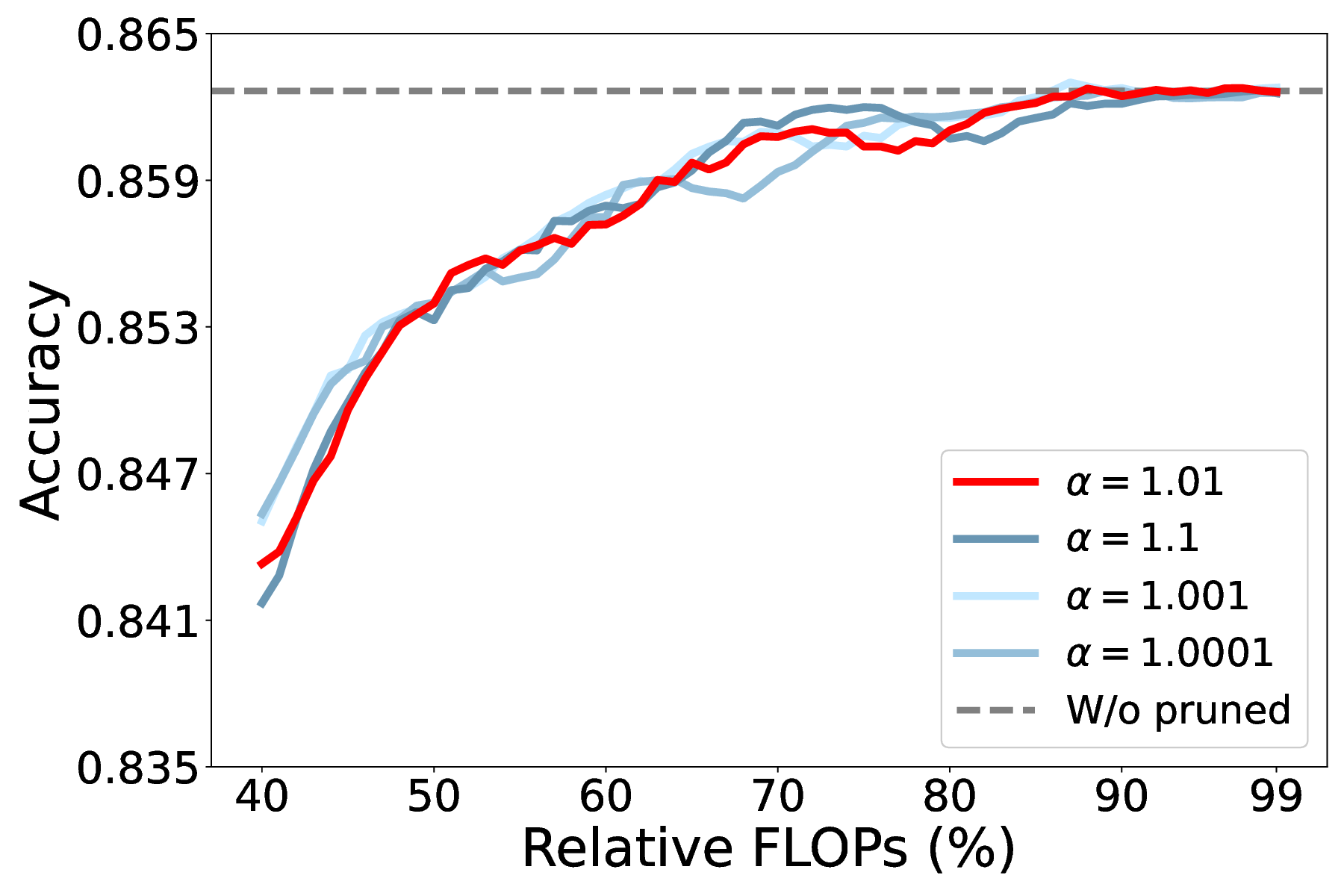}
}

\caption{The value of $\alpha = 1.01$ demonstrates exceptional performance in highly pruned models compared to other $\alpha$ values utilized in R\'enyi entropy estimation.}  
\label{fig:compare_on_alpha}    
\end{figure}

In this series of experiments, we would like to find out what could be the best choice of $\alpha$,
for the estimation of R\'enyi's $\alpha$-order Entropy.
As shown in FIG.~\ref{fig:compare_on_alpha},
choosing $\alpha =1.01$ gives the best result overall for five different tasks.

\subsubsection{Sample Number for MDS}

\begin{figure}[tb]
\centering 
\includegraphics[width=0.50\textwidth]{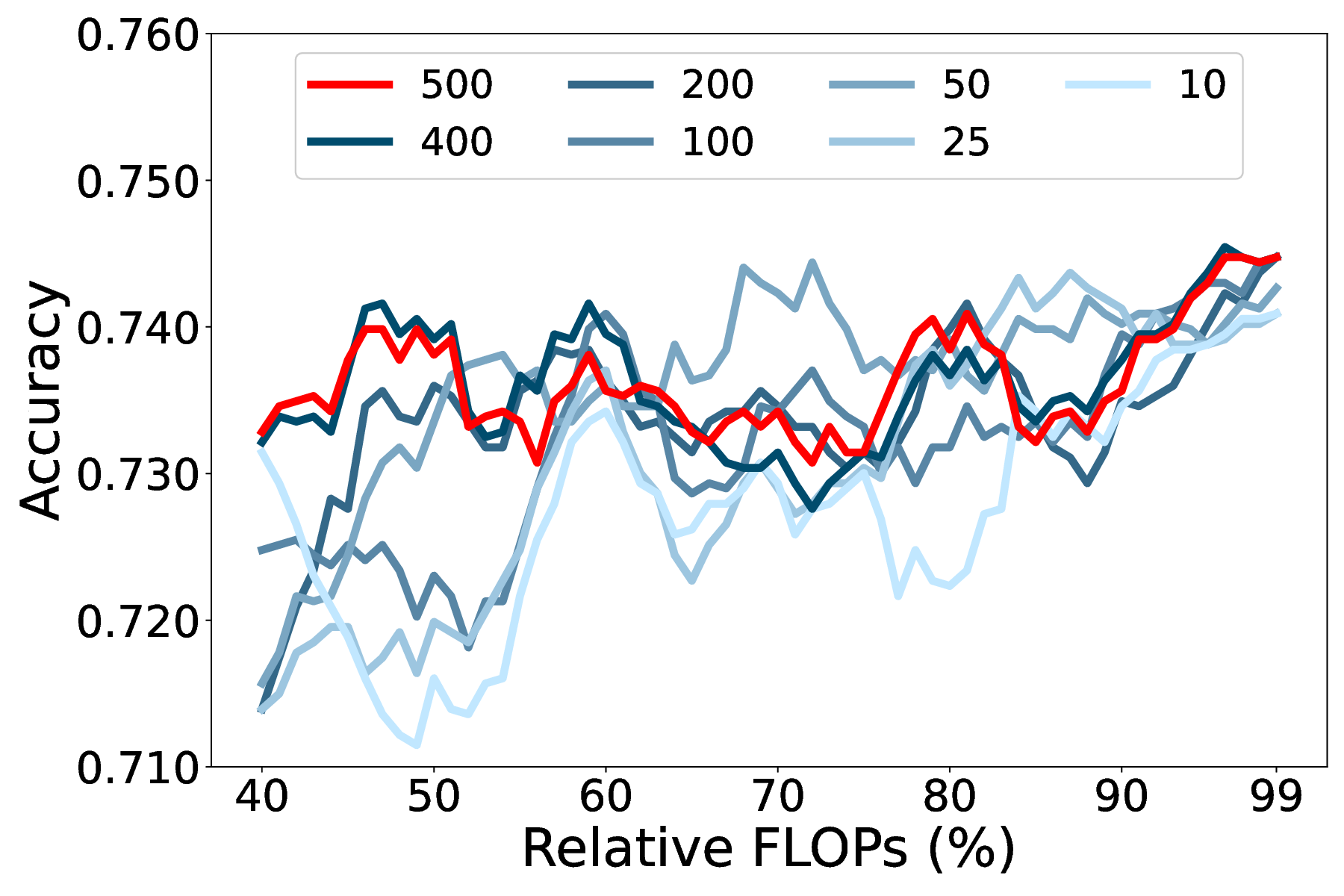}
\caption{In the MRPC experiment, the performance converges when
the number of samples exceeds $400$.}
\label{fig:compare_on_random_samples}    
\end{figure}

The proposed procedure may not always produce the same unique compressed model
because the solution to MDS and the selection of representatives
may not remain the same. To deal with that, we sample a few random seeds
and choose the best one based on Eq.~\ref{eq:kl}.
The result from comparing different numbers of samples
is shown in Fig.~\ref{fig:compare_on_random_samples}.

%

\section{Conclusion}
\label{sec:conc}

In this work, we proposed a novel method for large-scale language model pruning.
The pruning method is based on a precise estimation
of the mutual information on the representation neurons.
Moreover, a matrix-based R\'{e}nyi's $\alpha$-order entropy 
is adopted to realize such estimation.
In the estimation, we proposed a strategy to
find out the kernel width parameter, more effective than previous approaches.
Overall, the pruning method needs no label information nor retraining
to complete the task. Through a few series of experiments, we can conclude that
the proposed method indeed produced the pruned model,
more effective, but smaller than those from other state-of-the-art pruning approaches.
The superiority covers almost all unsupervised approaches
and a few supervised approaches. The result is also similar to that offered
by self-supervised learning. In the future, we would like to take an even
larger model than what we can process now
to confirm the scalability of the proposed method.

\balance

\bibliographystyle{unsrt}

\bibliography{BIB/IBref,BIB/ml,BIB/cs,BIB/tm}

\begin{thebibliography}{10}

\bibitem{touvron2023llama}
Hugo Touvron, Louis Martin, Kevin Stone, Peter Albert, Amjad Almahairi, Yasmine
  Babaei, Nikolay Bashlykov, Soumya Batra, Prajjwal Bhargava, Shruti Bhosale,
  et~al.
\newblock Llama 2: Open foundation and fine-tuned chat models.
\newblock {\em arXiv preprint arXiv:2307.09288}, 2023.

\bibitem{anil2023palm}
Rohan Anil, Andrew~M Dai, Orhan Firat, Melvin Johnson, Dmitry Lepikhin,
  Alexandre Passos, Siamak Shakeri, Emanuel Taropa, Paige Bailey, Zhifeng Chen,
  et~al.
\newblock Palm 2 technical report.
\newblock {\em arXiv preprint arXiv:2305.10403}, 2023.

\bibitem{wu2024llama}
Chengyue Wu, Yukang Gan, Yixiao Ge, Zeyu Lu, Jiahao Wang, Ye~Feng, Ping Luo,
  and Ying Shan.
\newblock Llama pro: Progressive llama with block expansion.
\newblock {\em arXiv preprint arXiv:2401.02415}, 2024.

\bibitem{singhal2023towards}
Karan Singhal, Tao Tu, Juraj Gottweis, Rory Sayres, Ellery Wulczyn, Le~Hou,
  Kevin Clark, Stephen Pfohl, Heather Cole-Lewis, Darlene Neal, et~al.
\newblock Towards expert-level medical question answering with large language
  models.
\newblock {\em arXiv preprint arXiv:2305.09617}, 2023.

\bibitem{zhang2024tinyllama}
Peiyuan Zhang, Guangtao Zeng, Tianduo Wang, and Wei Lu.
\newblock Tinyllama: An open-source small language model.
\newblock {\em arXiv preprint arXiv:2401.02385}, 2024.

\bibitem{mitra2023orca}
Arindam Mitra, Luciano Del~Corro, Shweti Mahajan, Andres Codas, Clarisse
  Simoes, Sahaj Agarwal, Xuxi Chen, Anastasia Razdaibiedina, Erik Jones, Kriti
  Aggarwal, et~al.
\newblock Orca 2: Teaching small language models how to reason.
\newblock {\em arXiv preprint arXiv:2311.11045}, 2023.

\bibitem{reid2024gemini}
Machel Reid, Nikolay Savinov, Denis Teplyashin, Dmitry Lepikhin, Timothy
  Lillicrap, Jean-baptiste Alayrac, Radu Soricut, Angeliki Lazaridou, Orhan
  Firat, Julian Schrittwieser, et~al.
\newblock Gemini 1.5: Unlocking multimodal understanding across millions of
  tokens of context.
\newblock {\em arXiv preprint arXiv:2403.05530}, 2024.

\bibitem{team2024gemma}
Gemma Team, Thomas Mesnard, Cassidy Hardin, Robert Dadashi, Surya Bhupatiraju,
  Shreya Pathak, Laurent Sifre, Morgane Rivi{\`e}re, Mihir~Sanjay Kale,
  Juliette Love, et~al.
\newblock Gemma: Open models based on gemini research and technology.
\newblock {\em arXiv preprint arXiv:2403.08295}, 2024.

\bibitem{zheng2024InternLM2}
Zheng Cai, Maosong Cao, and Haojiong~Chen et~al.
\newblock Internlm2 technical report.
\newblock {\em arXiv:2403.17297}, 2024.

\bibitem{achiam2023gpt}
Josh Achiam, Steven Adler, Sandhini Agarwal, Lama Ahmad, Ilge Akkaya,
  Florencia~Leoni Aleman, Diogo Almeida, Janko Altenschmidt, Sam Altman,
  Shyamal Anadkat, et~al.
\newblock Gpt-4 technical report.
\newblock {\em arXiv preprint arXiv:2303.08774}, 2023.

\bibitem{tishby2000information}
Naftali Tishby, Fernando~C Pereira, and William Bialek.
\newblock The information bottleneck method.
\newblock {\em arXiv preprint physics/0004057}, 2000.

\bibitem{tishby2015deep}
Naftali Tishby and Noga Zaslavsky.
\newblock Deep learning and the information bottleneck principle.
\newblock In {\em 2015 IEEE Information Theory Workshop (itw)}, pages 1--5.
  IEEE, 2015.

\bibitem{shwartz2017opening}
Ravid Shwartz-Ziv and Naftali Tishby.
\newblock Opening the black box of deep neural networks via information.
\newblock {\em arXiv preprint arXiv:1703.00810}, 2017.

\bibitem{lecun1989optimal}
Yann LeCun, John Denker, and Sara Solla.
\newblock Optimal brain damage.
\newblock {\em Advances in Neural Information Processing Systems}, 2, 1989.

\bibitem{han2015learning}
Song Han, Jeff Pool, John Tran, and William Dally.
\newblock Learning both weights and connections for efficient neural network.
\newblock {\em Advances in Neural Information Processing Systems}, 28, 2015.

\bibitem{li2016pruning}
Hao Li, Asim Kadav, Igor Durdanovic, Hanan Samet, and Hans~Peter Graf.
\newblock Pruning filters for efficient convnets.
\newblock {\em arXiv preprint arXiv:1608.08710}, 2016.

\bibitem{ma2023llm}
Xinyin Ma, Gongfan Fang, and Xinchao Wang.
\newblock Llm-pruner: On the structural pruning of large language models.
\newblock {\em Advances in Neural Information Processing Systems},
  36:21702--21720, 2023.

\bibitem{frantar2023massive}
Elias Frantar and Dan Alistarh.
\newblock Massive language models can be accurately pruned in one-shot.
\newblock {\em arXiv preprint arXiv:2301.00774}, 2023.

\bibitem{hinton2015distilling}
Geoffrey Hinton, Oriol Vinyals, and Jeff Dean.
\newblock Distilling the knowledge in a neural network.
\newblock {\em arXiv preprint arXiv:1503.02531}, 2015.

\bibitem{tung2019similarity}
Frederick Tung and Greg Mori.
\newblock Similarity-preserving knowledge distillation.
\newblock In {\em Proceedings of the IEEE/CVF International Conference on
  Computer Vision}, pages 1365--1374, 2019.

\bibitem{gu2023knowledge}
Yuxian Gu, Li~Dong, Furu Wei, and Minlie Huang.
\newblock Knowledge distillation of large language models.
\newblock {\em arXiv preprint arXiv:2306.08543}, 2023.

\bibitem{liu2021post}
Zhenhua Liu, Yunhe Wang, Kai Han, Wei Zhang, Siwei Ma, and Wen Gao.
\newblock Post-training quantization for vision transformer.
\newblock {\em Advances in Neural Information Processing Systems},
  34:28092--28103, 2021.

\bibitem{liu2023llm}
Zechun Liu, Barlas Oguz, Changsheng Zhao, Ernie Chang, Pierre Stock, Yashar
  Mehdad, Yangyang Shi, Raghuraman Krishnamoorthi, and Vikas Chandra.
\newblock Llm-qat: Data-free quantization aware training for large language
  models.
\newblock {\em arXiv preprint arXiv:2305.17888}, 2023.

\bibitem{ding20224}
Shaojin Ding, Phoenix Meadowlark, Yanzhang He, Lukasz Lew, Shivani Agrawal, and
  Oleg Rybakov.
\newblock 4-bit conformer with native quantization aware training for speech
  recognition.
\newblock {\em arXiv preprint arXiv:2203.15952}, 2022.

\bibitem{povey2018semi}
Daniel Povey, Gaofeng Cheng, Yiming Wang, Ke~Li, Hainan Xu, Mahsa Yarmohammadi,
  and Sanjeev Khudanpur.
\newblock Semi-orthogonal low-rank matrix factorization for deep neural
  networks.
\newblock In {\em Interspeech}, pages 3743--3747, 2018.

\bibitem{wu2023zeroquant}
Xiaoxia Wu, Zhewei Yao, and Yuxiong He.
\newblock Zeroquant-fp: A leap forward in llms post-training w4a8 quantization
  using floating-point formats.
\newblock {\em arXiv preprint arXiv:2307.09782}, 2023.

\bibitem{blalock2020state}
Davis Blalock, Jose~Javier Gonzalez~Ortiz, Jonathan Frankle, and John Guttag.
\newblock What is the state of neural network pruning?
\newblock {\em Proceedings of machine learning and systems}, 2:129--146, 2020.

\bibitem{zhu2023survey}
Xunyu Zhu, Jian Li, Yong Liu, Can Ma, and Weiping Wang.
\newblock A survey on model compression for large language models.
\newblock {\em arXiv preprint arXiv:2308.07633}, 2023.

\bibitem{zhang2023pruning}
Mingyang Zhang, Chunhua Shen, Zhen Yang, Linlin Ou, Xinyi Yu, Bohan Zhuang,
  et~al.
\newblock Pruning meets low-rank parameter-efficient fine-tuning.
\newblock {\em arXiv preprint arXiv:2305.18403}, 2023.

\bibitem{nova2023gradient}
Azade Nova, Hanjun Dai, and Dale Schuurmans.
\newblock Gradient-free structured pruning with unlabeled data.
\newblock {\em arXiv preprint arXiv:2303.04185}, 2023.

\bibitem{ganesh2021compressing}
Prakhar Ganesh, Yao Chen, Xin Lou, Mohammad~Ali Khan, Yin Yang, Hassan Sajjad,
  Preslav Nakov, Deming Chen, and Marianne Winslett.
\newblock Compressing large-scale transformer-based models: A case study on
  bert.
\newblock {\em Transactions of the Association for Computational Linguistics},
  9:1061--1080, 2021.

\bibitem{cover1999elements}
Thomas~M Cover.
\newblock {\em Elements of information theory}.
\newblock John Wiley \& Sons, 1999.

\bibitem{wickstrom2019information}
Kristoffer Wickstr{\o}m, Sigurd L{\o}kse, Michael Kampffmeyer, Shujian Yu, Jose
  Principe, and Robert Jenssen.
\newblock Information plane analysis of deep neural networks via matrix-based
  renyi's entropy and tensor kernels.
\newblock {\em arXiv preprint arXiv:1909.11396}, 2019.

\bibitem{giraldo2014measures}
Luis~Gonzalo Sanchez~Giraldo, Murali Rao, and Jose~C Principe.
\newblock Measures of entropy from data using infinitely divisible kernels.
\newblock {\em IEEE Transactions on Information Theory}, 61(1):535--548, 2014.

\bibitem{yu2019multivariate}
Shujian Yu, Luis Gonzalo~Sanchez Giraldo, Robert Jenssen, and Jose~C Principe.
\newblock Multivariate extension of matrix-based r{\'e}nyi's $\alpha$-order
  entropy functional.
\newblock {\em IEEE Transactions on Pattern Analysis and Machine Intelligence},
  42(11):2960--2966, 2019.

\bibitem{voita2019analyzing}
Elena Voita, David Talbot, Fedor Moiseev, Rico Sennrich, and Ivan Titov.
\newblock Analyzing multi-head self-attention: Specialized heads do the heavy
  lifting, the rest can be pruned.
\newblock {\em arXiv preprint arXiv:1905.09418}, 2019.

\bibitem{liu2021ebert}
Zejian Liu, Fanrong Li, Gang Li, and Jian Cheng.
\newblock Ebert: Efficient bert inference with dynamic structured pruning.
\newblock In {\em Findings of the Association for Computational Linguistics:
  ACL-IJCNLP 2021}, pages 4814--4823, 2021.

\bibitem{kwon2022fast}
Woosuk Kwon, Sehoon Kim, Michael~W Mahoney, Joseph Hassoun, Kurt Keutzer, and
  Amir Gholami.
\newblock A fast post-training pruning framework for transformers.
\newblock {\em Advances in Neural Information Processing Systems},
  35:24101--24116, 2022.

\bibitem{yang-etal-2022-textpruner}
Ziqing Yang, Yiming Cui, and Zhigang Chen.
\newblock {T}ext{P}runer: A model pruning toolkit for pre-trained language
  models.
\newblock In {\em Proceedings of the 60th Annual Meeting of the Association for
  Computational Linguistics: System Demonstrations}, pages 35--43, Dublin,
  Ireland, May 2022. Association for Computational Linguistics.

\bibitem{park2023accurate}
Seungcheol Park, Hojun Choi, and U~Kang.
\newblock Accurate retraining-free pruning for pretrained encoder-based
  language models.
\newblock In {\em The Twelfth International Conference on Learning
  Representations}, 2023.

\bibitem{ma2024llm}
Xinyin Ma, Gongfan Fang, and Xinchao Wang.
\newblock Llm-pruner: On the structural pruning of large language models.
\newblock {\em Advances in Neural Information Processing Systems}, 36, 2024.

\bibitem{lagunas2021block}
Fran{\c{c}}ois Lagunas, Ella Charlaix, Victor Sanh, and Alexander~M Rush.
\newblock Block pruning for faster transformers.
\newblock {\em arXiv preprint arXiv:2109.04838}, 2021.

\bibitem{an2023fluctuation}
Yongqi An, Xu~Zhao, Tao Yu, Ming Tang, and Jinqiao Wang.
\newblock Fluctuation-based adaptive structured pruning for large language
  models.
\newblock {\em arXiv preprint arXiv:2312.11983}, 2023.

\bibitem{zhou2022feature}
Hongfang Zhou, Xiqian Wang, and Rourou Zhu.
\newblock Feature selection based on mutual information with correlation
  coefficient.
\newblock {\em Applied Intelligence}, pages 1--18, 2022.

\bibitem{lewis1992feature}
David~D Lewis.
\newblock Feature selection and feature extraction for text categorization.
\newblock In {\em Speech and Natural Language: Proceedings of a Workshop Held
  at Harriman, New York, February 23-26, 1992}, 1992.

\bibitem{peng2005feature}
Hanchuan Peng, Fuhui Long, and Chris Ding.
\newblock Feature selection based on mutual information criteria of
  max-dependency, max-relevance, and min-redundancy.
\newblock {\em IEEE Transactions on Pattern Analysis and Machine Intelligence},
  27(8):1226--1238, 2005.

\bibitem{fan2021layer}
Chun Fan, Jiwei Li, Xiang Ao, Fei Wu, Yuxian Meng, and Xiaofei Sun.
\newblock Layer-wise model pruning based on mutual information.
\newblock {\em arXiv preprint arXiv:2108.12594}, 2021.

\bibitem{sarvani2022hrel}
CH~Sarvani, Mrinmoy Ghorai, Shiv~Ram Dubey, and SH~Shabbeer Basha.
\newblock Hrel: Filter pruning based on high relevance between activation maps
  and class labels.
\newblock {\em Neural Networks}, 147:186--197, 2022.

\bibitem{lee2021information}
Sungyeop Lee and Junghyo Jo.
\newblock Information flows of diverse autoencoders.
\newblock {\em arXiv preprint arXiv:2102.07402}, 2021.

\bibitem{holmes2019estimation}
Caroline~M Holmes and Ilya Nemenman.
\newblock Estimation of mutual information for real-valued data with error bars
  and controlled bias.
\newblock {\em Physical Review E}, 100(2):022404, 2019.

\bibitem{kraskov2004estimating}
Alexander Kraskov, Harald St{\"o}gbauer, and Peter Grassberger.
\newblock Estimating mutual information.
\newblock {\em Physical review E}, 69(6):066138, 2004.

\bibitem{belghazi2018mutual}
Mohamed~Ishmael Belghazi, Aristide Baratin, Sai Rajeshwar, Sherjil Ozair,
  Yoshua Bengio, Aaron Courville, and Devon Hjelm.
\newblock Mutual information neural estimation.
\newblock In {\em International Conference on Machine Learning}, pages
  531--540. PMLR, 2018.

\bibitem{scott2015multivariate}
David~W Scott.
\newblock {\em Multivariate density estimation: theory, practice, and
  visualization}.
\newblock John Wiley \& Sons, 2015.

\bibitem{cox2000multidimensional}
Trevor~F Cox and Michael~AA Cox.
\newblock {\em Multidimensional scaling}.
\newblock CRC press, 2000.

\bibitem{cristianini2001kernel}
Nello Cristianini, John Shawe-Taylor, Andre Elisseeff, and Jaz Kandola.
\newblock On kernel-target alignment.
\newblock {\em Advances in Neural Information Processing Systems}, 14, 2001.

\bibitem{khaki2024the}
Samir Khaki and Konstantinos~N Plataniotis.
\newblock The need for speed: Pruning transformers with one recipe.
\newblock In {\em The Twelfth International Conference on Learning
  Representations}, 2024.

\bibitem{turc1908well}
Iulia Turc, Ming-Wei Chang, Kenton Lee, and Kristina Toutanova.
\newblock Well-read students learn better: On the importance of pre-training
  compact models.
\newblock {\em arXiv preprint arXiv:1908.08962}, 1908.

\bibitem{wang2018glue}
Alex Wang, Amanpreet Singh, Julian Michael, Felix Hill, Omer Levy, and Samuel~R
  Bowman.
\newblock Glue: A multi-task benchmark and analysis platform for natural
  language understanding.
\newblock {\em arXiv preprint arXiv:1804.07461}, 2018.

\bibitem{socher2013recursive}
Richard Socher, Alex Perelygin, Jean Wu, Jason Chuang, Christopher~D Manning,
  Andrew~Y Ng, and Christopher Potts.
\newblock Recursive deep models for semantic compositionality over a sentiment
  treebank.
\newblock In {\em Proceedings of the 2013 Conference on Empirical Methods in
  Natural Language Processing}, pages 1631--1642, 2013.

\bibitem{cer2017semeval}
Daniel Cer, Mona Diab, Eneko Agirre, Inigo Lopez-Gazpio, and Lucia Specia.
\newblock Semeval-2017 task 1: Semantic textual similarity-multilingual and
  cross-lingual focused evaluation.
\newblock {\em arXiv preprint arXiv:1708.00055}, 2017.

\bibitem{dolan2005automatically}
Bill Dolan and Chris Brockett.
\newblock Automatically constructing a corpus of sentential paraphrases.
\newblock In {\em Third International Workshop on Paraphrasing (IWP2005)}. Asia
  Federation of Natural Language Processing, January 2005.

\bibitem{loshchilov2018decoupled}
Ilya Loshchilov and Frank Hutter.
\newblock Decoupled weight decay regularization.
\newblock In {\em International Conference on Learning Representations}, 2019.

\end{thebibliography}

\end{document}